\documentclass[lettersize,journal]{IEEEtran}
\usepackage[colorlinks,linkcolor=red,citecolor=blue]{hyperref}
\usepackage[caption=false,font=normalsize,labelfont=sf,textfont=sf]{subfig}
\usepackage{amsmath,amsfonts}
\usepackage{algorithmic}
\usepackage{algorithm}
\usepackage{array}
\usepackage{textcomp}
\usepackage{stfloats}
\usepackage{url}
\usepackage{verbatim}
\usepackage{graphicx}
\usepackage{cite}
\usepackage{lineno}
\usepackage{amssymb}
\usepackage{booktabs}
\usepackage{caption}
\usepackage{cuted}
\usepackage{pifont}
\usepackage{etoolbox,lipsum}
\usepackage{soul}
\usepackage{booktabs}
\usepackage{multirow}
\usepackage{enumitem}
\newcommand{\revision}[1]{\textcolor{black}{#1}}
\newcommand{\secondrevision}[1]{\textcolor{black}{#1}}

\begin{document}

\title{SuperCarver: Texture-Consistent 3D Geometry Super-Resolution for High-Fidelity \\ Surface Detail Generation}

\author{IEEE Publication Technology,~\IEEEmembership{Staff,~IEEE,}
	\thanks{This paper was produced by the IEEE Publication Technology Group. They are in Piscataway, NJ.}
}

\author{Qijian Zhang, Xiaozheng Jian, Xuan Zhang, Wenping Wang, Junhui Hou
  \thanks{This work was supported in part by the  Natural Science Foundation of China under Grant 62422118, and in part by the Hong Kong Research Grants Council under Grants 11219324, 11219422, and N\_CityU1114/25. (\textit{Corresponding author: Junhui Hou})}
	\thanks{Q. Zhang, X. Jian, and X. Zhang are with Tencent Games, China. E-mail: keegan.zqj@gmail.com; ajian@tencent.com; libzhang@tencent.com.}
	\thanks{W. Wang is with the Department of Computer Science \& Engineering, Texas A \& M University, USA. Email: wenping@tamu.edu.}
	\thanks{J. Hou is with the Department of Computer Science, City University of Hong Kong, Hong Kong SAR, China. Email: jh.hou@cityu.edu.hk.}
}

\maketitle

\begin{abstract}
    Conventional production workflow of high-precision mesh assets necessitates a cumbersome and laborious process of manual sculpting by specialized 3D artists/modelers. The recent years have witnessed remarkable advances in AI-empowered 3D content creation for generating plausible structures and intricate appearances from images or text prompts. However, synthesizing realistic surface details still poses great challenges, and enhancing the geometry fidelity of existing lower-quality 3D meshes (instead of image/text-to-3D generation) remains an open problem. In this paper, we introduce SuperCarver, a 3D geometry super-resolution pipeline for supplementing texture-consistent surface details onto a given coarse mesh. We start by rendering the original textured mesh into the image domain from multiple viewpoints. To achieve detail boosting, we construct a deterministic prior-guided normal diffusion model, which is fine-tuned on a carefully curated dataset of paired detail-lacking and detail-rich normal map renderings. To update mesh surfaces from potentially imperfect normal map predictions, we design a noise-resistant inverse rendering scheme through deformable distance field. Experiments demonstrate that our SuperCarver is capable of generating realistic and expressive surface details depicted by the actual texture appearance, making it a powerful tool to both upgrade historical low-quality 3D assets and reduce the workload of sculpting high-poly meshes.
\end{abstract}


\section{Introduction} \label{sec:introduction}

3D modeling plays a critical role in various industrial design and digital entertainment fields, such as movies, animations, video games, robotics, and virtual/augmented reality. In practice, the geometric accuracy of mesh assets directly influence the expressiveness of visual presentation and the performance of specific downstream tasks. Nevertheless, the production of high-quality mesh models heavily depends on manual efforts of specialized 3D modelers, which can be highly cumbersome and labor-intensive.

Generally, the complete industrial 3D modeling pipeline can be broadly divided into three sequential stages in terms of the geometric complexity of mesh surfaces:
\begin{itemize}[leftmargin=23.5pt, topsep=1.0pt]
    \item[1)] \textbf{Mid-Poly:} This stage primarily aims at converting initial concept design works (typically 2D sketch drawings) to 3D meshes, roughly reproducing the designer's vision in terms of the overall style and spatial structure.
    \item[2)] \textbf{High-Poly:} To produce high-quality geometric surfaces, the preceding mid-poly meshes need to be carved within specialized digital sculpting software (typically ZBrush) to obtain high-poly meshes with realistic and expressive surface details.
    \item[3)] \textbf{Low-Poly:} To satisfy efficiency requirements of varying tasks and devices, high-poly meshes need to be significantly simplified to newly create low-poly meshes with typically thousands or hundreds of polygons, followed by UV unwrapping, detail (normal/bump map) baking, and texture painting. 
\end{itemize}

Among the above three modeling stages, the effects of high-poly surface sculpting dominate the ultimate graphics quality and visual realism, while accounting for the majority of the overall workload. Therefore, there is an urgent need to develop automated surface detail generation techniques to accelerate the production of high-fidelity mesh production.

The recent advancements of deep learning-based generative modeling have led to remarkable strides in 3D content generation \cite{li2024advances}. Early zero-shot architectures such as DreamFields~\cite{jain2022zero} and DreamFusion~\cite{poole2023dreamfusion} distill prior knowledge from pre-trained vision-language multi-modal models \cite{radford2021learning} and/or 2D image diffusion models \cite{rombach2022high}. Nevertheless, such per-shape optimization process can typically suffer from inefficiency and inter-view inconsistency problems. Since the emergence of large-scale mesh datasets \cite{deitke2023objaverse,deitke2023objaversexl}, feed-forward 3D generation frameworks such as LRM~\cite{hong2024lrm} and InstantMesh~\cite{xu2024instantmesh} receive increasing attention, which achieve end-to-end training and finish inference within several seconds. To further enhance generation quality without significantly compromising efficiency, another family of research such as SyncDreamer~\cite{liu2024syncdreamer}, Wonder3D~\cite{long2024wonder3d}, and Unique3D~\cite{wu2024unique3d} proposes to apply multi-view diffusion models to synthesize 2D novel views and then reconstruct the target 3D mesh through iterative optimization procedures. Moreover, native 3D latent diffusion architectures such as CLAY~\cite{zhang2024clay} also demonstrate tremendous potential. The latest state-of-the-arts \cite{ye2025hi3dgen,wu2025direct3d,li2025sparc3d,lai2025hunyuan3d} have continuously advanced the geometric accuracy of the generated meshes to new heights.

Functionally, previous 3D generation frameworks are particularly designed for creating \textbf{new 3D assets} according to low-dimensional textual (1D-to-3D) or visual (2D-to-3D) prompts. However, there still exists a large amount of practical demands to upgrade the geometric fidelity of \textbf{existing 3D assets} while strictly preserving the original structural characteristics, which is a different mesh-to-mesh (3D-to-3D) task scenario. \revision{Besides, there exist several lines of relevant studies, such as point cloud upsampling \cite{yu2018pu,qian2020pugeo}, reference-based \cite{chen2024decollage} or text-guided \cite{raj2023dreambooth3d} 3D editing, dense geometric map (i.e., depth/normal) \cite{bae2024rethinking,ke2024repurposing,fu2024geowizard} or PBR material \cite{wang2024boosting} estimation, but none of them are suitable for our targeted surface detail generation. More recent works \cite{deng2024detailgen3d} devote to 3D geometry enhancement, but the detailization performance and the identity-preserving capability are still far from satisfactory.}

\begin{figure*}[t!]
    \centering
    \includegraphics[width=0.8\linewidth]{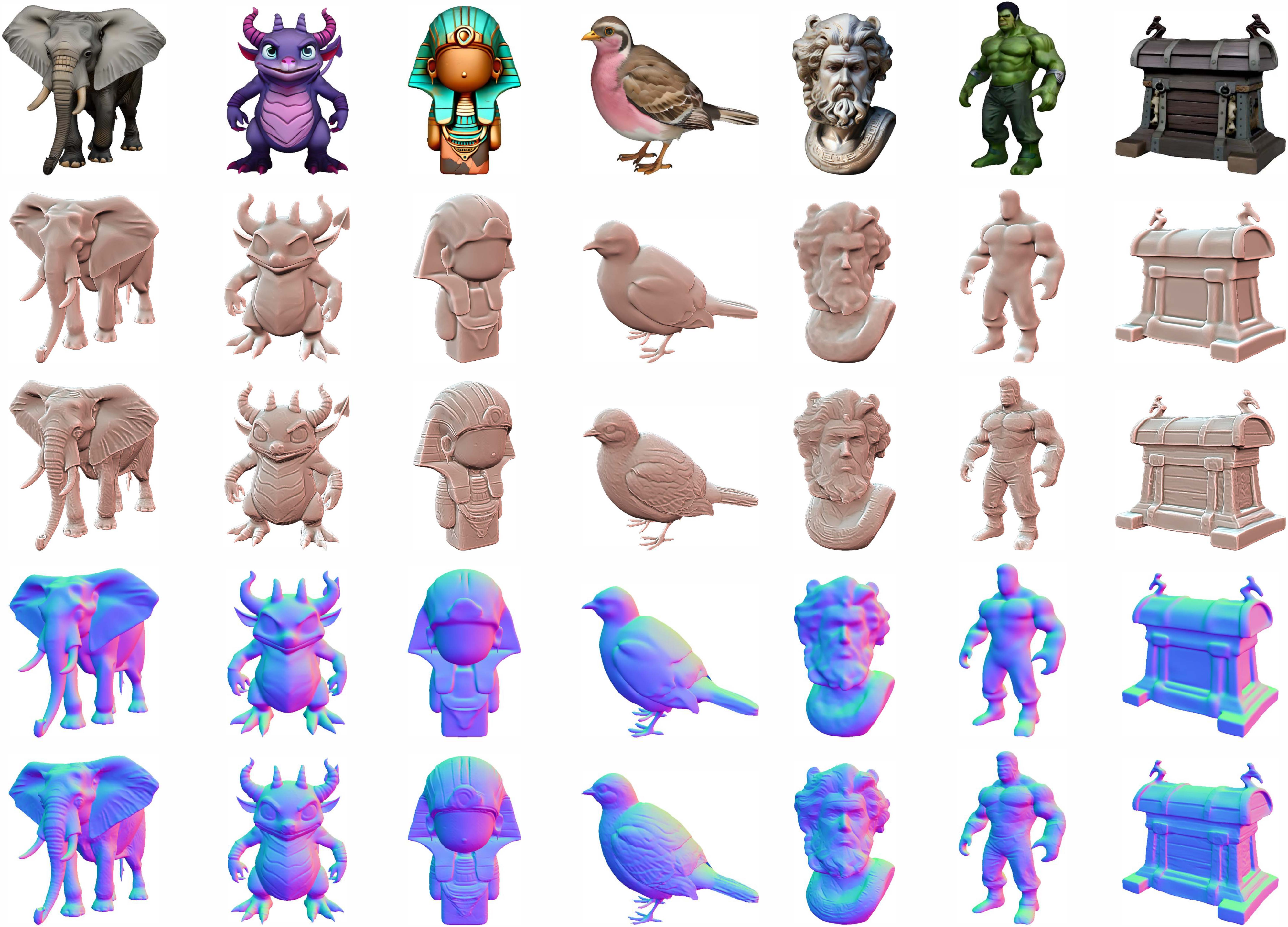}
    \caption{Given an input 3D textured model lacking geometric details, \textbf{SuperCarver} generates realistic and expressive 3D surface details consistent with its texture appearance within 1\,$\sim$\,2 minutes, significantly enhancing the production efficiency of high-quality mesh assets. (\textit{Row 1}: The input 3D textured meshes. \textit{Rows 2\&4}: The original coarse surfaces and normal maps. \textit{Rows 3\&5}: The resulting high-fidelity mesh surfaces and normal maps after our geometry super-resolution.)}
    \label{fig:teaser}
    \vspace{-0.5cm}
\end{figure*}

\revision{This paper introduces SuperCarver, a two-stage processing framework for geometry super-resolution operating on given 3D textured meshes to generate realistic and expressive surface details that are faithful to the corresponding texture appearances, as shown in Figure~\ref{fig:teaser}.}

Considering the difficulties of directly learning 3D structural details, we choose normal maps, a 2.5D representation format, to efficiently bridge 2D and 3D domains. Accordingly, we start by rendering the input textured mesh into image space, which facilitates leveraging the power and pre-trained knowledge of 2D image diffusion architectures. Still, accurately and robustly predicting normal maps through diffusion faces several challenges, as outlined below: 
\begin{itemize}[leftmargin=23.5pt, topsep=1.0pt]
    \item[1)] Normals are a form of sensitive and vulnerable geometric attributes. Moderate numerical perturbations can lead to obvious distortion and damage to the underlying surface. Without sufficient priors, normals predicted in the image space often fail to align with the actual surface structures of the given meshes.
    \item[2)] Due to the inherent stochasticity of diffusion models, it is inevitable that normal predictions exhibit certain degrees of jittering across multiple times of inferences.
\end{itemize}

To address these problems, inspired by previous attempts to stabilize \cite{xu2025what} and accelerate \cite{yue2023resshift} image diffusion mechanisms, we adopt a chain of deterministic interpolations from rendered detail-lacking and detail-rich normal maps. In the meantime, necessary conditioning signals are incorporated by extracting both semantic and geometric priors from rendered color images and depth maps.

Having enhanced the detail richness of normal maps from multiple viewpoints, it is straightforward to perform gradient-based mesh updating through the optimization loop of inverse rendering. However, optimizing the mesh surface from potentially imperfect multi-view normal map predictions is prone to causing unruly surface degradation. Therefore, we customize a simple yet effective noise-resistant inverse rendering approach based on distance field deformation to deduce the corresponding surface details.

Through comprehensive qualitative illustrations and quantitative evaluations, we demonstrate that our SuperCarver framework can supplement highly-expressive and texture-consistent surface details to various textured meshes with unsatisfactory geometric quality. The main contributions and superiorities of our work can be summarized as follows: 
\begin{itemize}[leftmargin=23.5pt, topsep=1.0pt]
    \item[1)] \revision{We introduce a novel mesh-to-mesh 3D geometry super-resolution learning framework.} Our targeted task differs from previous 1D/2D-to-3D generative paradigms, offering new capabilities of upgrading the geometric quality of given mesh models in a non-destructive and identity-preserving manner. 
    \item[2)] We propose a two-stage normal-centric pipeline featured by deterministic diffusion mechanism and noise-resistant distance field deformation. 
    \item[3)] Our approach fills the gap in the current 3D generative modeling community for high-poly surface sculpting and historical low-quality mesh asset upgrading.
\end{itemize}

The remainder of this paper is organized as follows. Section~\ref{sec:related-work} systematically reviews different perspectives of research on 3D generation, optimization, and enhancement. Section~\ref{sec:method} introduces our SuperCarver framework for pixel-space normal boosting and geometry-space mesh updating. Section~\ref{sec:experiments} provides comprehensive evaluations, comparisons, and ablation studies to demonstrate the effectiveness of our approach. In the end, we conclude the whole paper in Section~\ref{sec:conclusion}.

\section{Related Work} \label{sec:related-work}

\subsection{3D Content Generation} \label{sec:rw:3d-gen}

\revision{3D content generation \cite{liu2025dreamreward,huang2025part,liu2025acc3d,ren2025neural,han2024super,zhang2022marching,metzer2023latent,wang2023score,zhang20233dshape2vecset,zhao2023michelangelo,chen2025dora} has always been a highly valuable yet rather challenging task.} Unfortunately, the limited data volume and the diversity of classic 3D shape repositories \cite{chang2015shapenet} largely restrict the advancement of this field. In order to bypass the reliance on large-scale 3D training data, zero-shot schemes \cite{jain2022zero,michel2022text2mesh} are designed to constrain the text-prompted 3D generation process using the off-the-shelf language-image prior of CLIP~\cite{radford2021learning}. \revision{With the advent of pre-trained 2D diffusion models \cite{rombach2022high}, score distillation \cite{poole2023dreamfusion} is investigated to inject text guidance \cite{wu2024consistent3d,miao2024dreamer,zhang2024text2nerf} into the optimization procedure of various neural field representations \cite{mildenhall2020nerf,kerbl20233d}.} Subsequent researches further make continuous advancements in improving generation quality and/or speed \cite{wang2024prolificdreamer,tang2024dreamgaussian}, enabling 3D editing \cite{raj2023dreambooth3d,sun2024dreamcraftd}, and supplementing image conditioning capabilities \cite{qian2024magic,shi2024mvdream}. Nevertheless, the distillation-based generation paradigm necessitates sophisticated processing to deal with inefficiency and inter-view inconsistency.

Thanks to the significant increase in the scale of 3D data \cite{deitke2023objaverse,deitke2023objaversexl}, it becomes promising to directly train feed-forward 3D generation networks. The pioneering approach of LRM~\cite{hong2024lrm} constructs an end-to-end regression framework to predict a triplane~\cite{chan2022efficient} NeRF representation, while Instant3D~\cite{li2024instant3d} develops a two-stage framework that begins with generating the ``view-compiled'' image grid from texts. Numerous follow-up studies continue to improve performances from different perspectives \cite{tochilkin2024triposr,li2025multi}, such as shifting to strong 3DGS~\cite{kerbl20233d} representations (LGM~\cite{tang2024lgm}), replacing transformers with convolutional architectures (CRM~\cite{wang2024crm}), and adding more geometric supervisions (InstantMesh~\cite{xu2024instantmesh}).

To further enhance the generation quality while maintaining satisfactory efficiency, viewpoint-conditioned 2D diffusion \cite{liu2023zero} is investigated to synthesize novel 2D views, followed by iterative optimization for sparse-view 3D reconstruction. However, lifting imperfect multi-view 2D predictions to 3D can be non-trivial and usually problematic. To ensure multi-view consistency, SyncDreamer~\cite{liu2024syncdreamer} designs synchronized multi-view noise prediction by constructing an intermediate 3D feature volume. In Zero123++~\cite{shi2023zero123++}, multiple images are tiled into a single 2D frame to model the joint distribution of all the pre-defined views. Wonder3D~\cite{long2024wonder3d} builds an effective multi-view cross-domain 2D diffusion pipeline controlled by a domain switcher to generate multi-view normal maps and color images, after which explicit 3D geometry is extracted via normal fusion. Era3D~\cite{li2024erad} incorporates explicit regression of camera parameters and designs more efficient and compact multi-view attention. Unique3D~\cite{wu2024unique3d} constructs a powerful image-to-3D framework with comprehensive improvements in multi-view generation and upscaling of color images and normal maps, together with an efficient multi-stage surface reconstruction algorithm.

Analogous to 2D diffusion for image generation, another promising direction is to develop native 3D diffusion architectures \cite{wu2024direct3d,zhang2024clay} for conditional 3D generation with different design choices of latent space representations. Additionally, auto-regressive generation of triangle sequences \cite{siddiqui2024meshgpt,chen2025meshanything} and UV-domain diffusion \cite{yan2024object} driven by regular geometry image representations \cite{gu2002geometry,zhang2022reggeonet,zhang2023flattening} also demonstrate new potential.

\subsection{Inverse Rendering and Mesh Optimization} \label{sec:rw:ir-mesh-opt}

Inverse rendering aims at reasoning the underlying physical attributes of a target 3D scene from its 2D image observations. Different 3D scene parameters, including geometry, textures, materials, and lighting, can be optimized through backpropagating the
gradients w.r.t. the 2D rendering outputs \cite{munkberg2022extracting,hasselgren2022shape}. Earlier research investigates various differentiable renderers \cite{kato2018neural,liu2019soft}. Later studies \cite{boss2021nerd,srinivasan2021nerv,zhang2021nerfactor} devote to learning neural field representations. For gradient-based mesh optimization, the most straightforward strategy is to explicitly perform surface deformation from pre-defined mesh templates \cite{wang2018pixel2mesh,Hanocka2020p2m}. Nevertheless, directly optimizing the mesh vertex positions can be prone to surface defect and degeneracy, thus requiring careful initialization, regularization, and remeshing \cite{khan2020surface,nicolet2021large,palfinger2022continuous}. Another choice is to employ occupancy and distance fields \cite{mescheder2019occupancy,chen2019learning,park2019deepsdf} for implicit surface modeling. Since traditional iso-surface extraction algorithms \cite{lorensen1987marching,nielson2004dual} cannot propagate gradients, various differentiable schemes \cite{chen2021neural,chen2022neural} are designed to operate on discrete occupancy grids (DMC~\cite{liao2018deep}), continuous signed distance fields (MeshSDF~\cite{remelli2020meshsdf}), and deformable tetrahedrons (DMTet~\cite{shen2021deep}). Moreover, NJF~\cite{aigerman2022neural} depicts the mappings between pairs of source and target triangles through Jacobian matrices. FlexiCubes~\cite{shen2023flexible} incorporates the carefully-chosen learnable parameters into \cite{nielson2004dual} for achieving much more flexible local adjustment of mesh vertices.

\begin{figure*}[t!]
    \centering
    \includegraphics[width=0.9\linewidth]{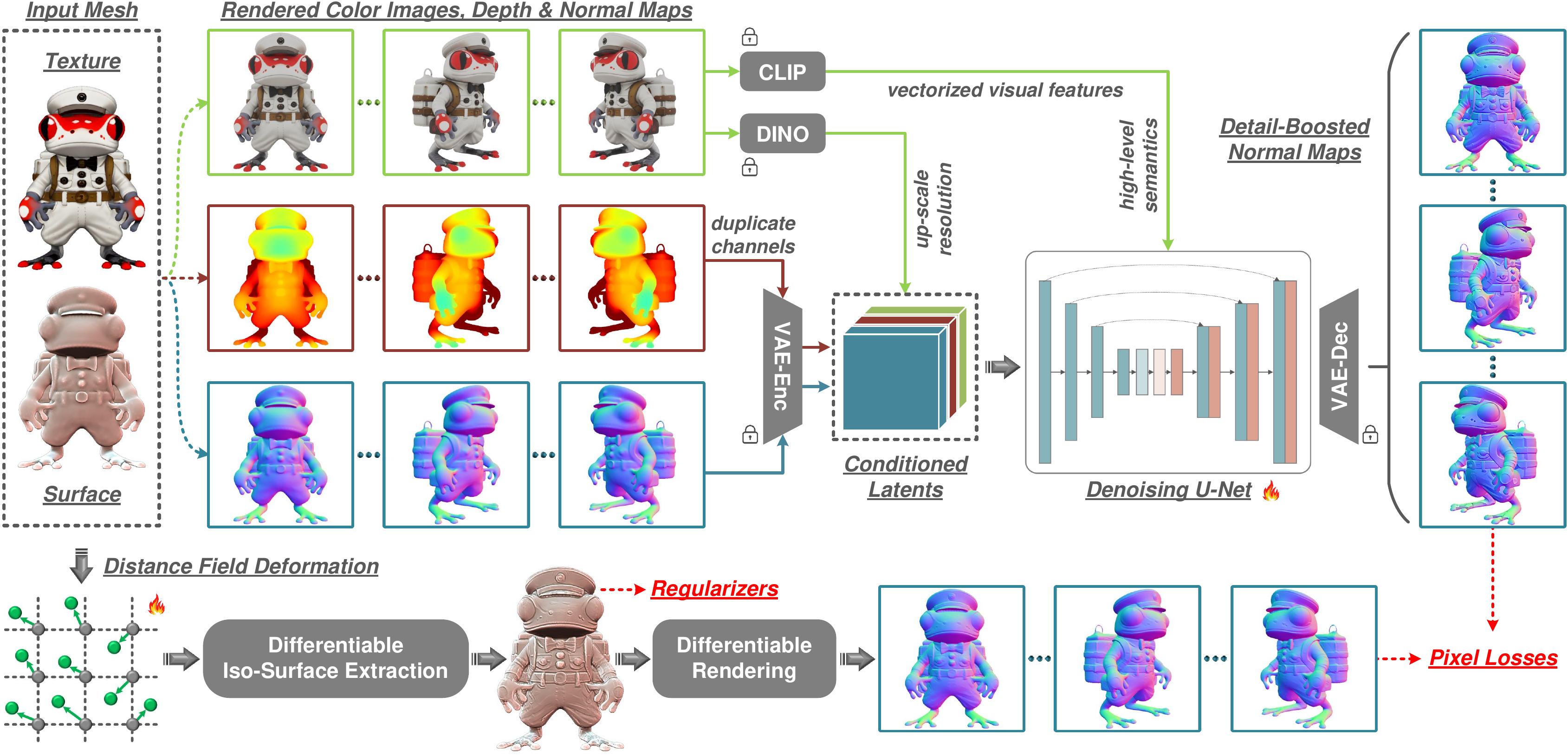}
    \caption{The overall processing pipeline of our proposed SuperCarver for texture-consistent geometry super-resolution. Given an input 3D textured model, we begin by rendering color, depth, and normal maps, which are fed into a customized deterministic prior-guided diffusion architecture to generate detail-boosted normal maps. To perform gradient-based mesh refinement from potentially imperfect multi-view normal map predictions, we further design a noise-resistant distance field deformation process to drive the optimization of high-fidelity 3D surface details.}
    \label{fig:overall-workflow}
    \vspace{-0.5cm}
\end{figure*}

\subsection{3D Geometry Enhancement} \label{sec:rw:geo-enhance}

Instead of directly generating textured meshes from images or text prompts, another line of research aims to enhance the geometric surface of given 3D models.

Similar to image super-resolution, 3D point cloud upsampling \cite{yu2018pu,qian2020pugeo,qian2021deep,mao2022pu} can produce a denser set of spatial points uniformly located on the underlying 3D surface from sparse and possibly noisy inputs. However, these approaches are rarely capable of generating new geometric details. When a high-quality source mesh is given as the reference, detail transfer \cite{berkiten2017learning} can synthesize similar geometric patterns on the target surface. More recent works \cite{chen2024decollage} further apply such an examplar-based paradigm on coarse voxels for few-shot 3D shape detailization. Text-guided mesh editing frameworks \cite{gao2023textdeformer} facilitate flexible and convenient geometry manipulation, but face challenges in accurate and fine-grained control. Additionally, PBR material generation \cite{wang2024boosting} typically involves estimating normal/bump maps, which can be potentially exploited for refinement purposes.

\revision{Moreover, Sherpa3D~\cite{liu2024sherpa3d} exploits a coarse 3D prior to guide a 2D diffusion model for mesh refinement. CraftsMan3D~\cite{li2025craftsman3d} incorporates a normal-driven mesh geometry refinement stage, in which new surface details can be hallucinated from coarse normal maps. DetailGen3D~\cite{deng2024detailgen3d} achieves geometry enhancement through training a rectified flow \cite{liu2023flow} model that maps the latent codes between coarse and fine geometries. Despite the current progress, it is still challenging to generate high-fidelity surface details and achieve geometry-texture consistency while effectively preserving the original structure.}

\section{Proposed Method} \label{sec:method}

\subsection{Overall Workflow}

For a 3D textured mesh model $\mathcal{M}$ with vertices $\mathcal{V}$ and faces $\mathcal{F}$, we aim at generating high-fidelity surface details reflected by its specific texture appearance, thus producing a high-poly version $\mathcal{M}^u = (\mathcal{V}^u, \mathcal{F}^u)$ of the input coarse mesh surface $\mathcal{M}$. As a geometry sculpting process, it is particularly required that the original spatial structures should be preserved.

Figure~\ref{fig:overall-workflow} illustrates the overall processing procedures of our proposed SuperCarver. Technically, we begin by rendering the input textured mesh $\mathcal{M}$ from a certain pre-specified viewpoint, respectively producing the corresponding 2D RGB color image $\mathcal{I} \in \mathbb{R}^{H \times W \times 3}$, normal map $\mathcal{N} \in \mathbb{R}^{H \times W \times 3}$, and depth map $\mathcal{D} \in \mathbb{R}^{H \times W}$. The subsequent 3D geometry super-resolution is achieved in two sequential stages. In the first stage, we propose a deterministic prior-guided diffusion architecture conditioned on diverse semantic and geometric cues to predict a detail-rich normal map $\mathcal{H} \in \mathbb{R}^{H \times W \times 3}$. Thus, by performing inferences from multiple different viewpoints, we can acquire a group of multi-view normal map predictions, denoted as $\{ \mathcal{H}_k \}_{k=1}^K$. In the second stage, we propose a noise-resistant inverse normal rendering approach based on deformable distance field representation to iteratively optimize mesh surface with geometric constraints. In what follows, we will detail the two stages and specifically explain our design choices.

\begin{figure*}[t!]
	\centering
	\includegraphics[width=0.8\linewidth]{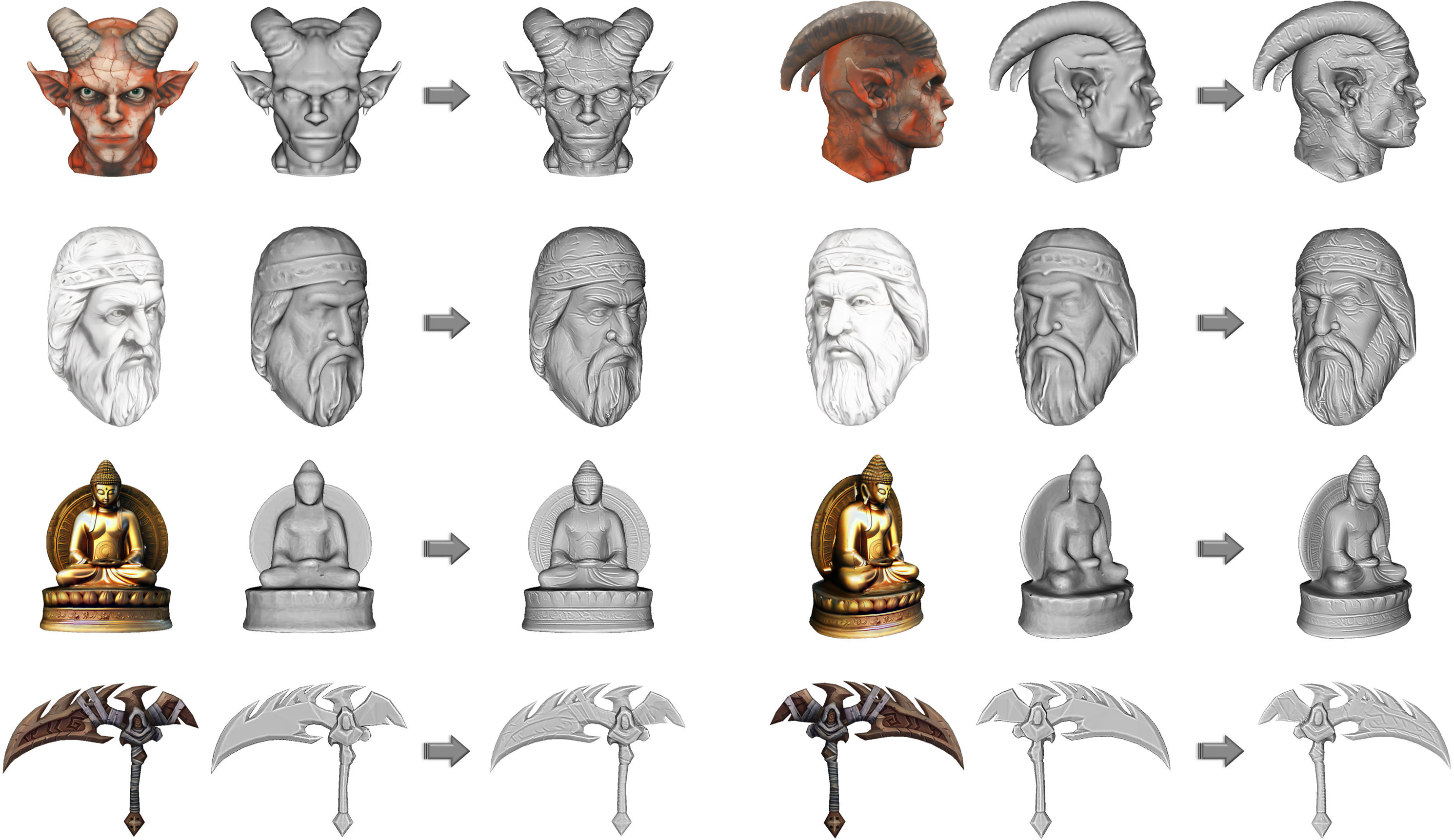}
	\caption{Results of applying our proposed SuperCarver for texture-consistent 3D geometry super-resolution.}
	\label{fig:more-results}
	\vspace{-0.5cm}
\end{figure*}

\subsection{Deterministic Prior-Guided Normal Diffusion} \label{sec:method:normal-diffusion}

\revision{Diffusion probabilistic models \cite{ho2020denoising} define a forward Markov chain gradually perturbing data into noise, and learn a reverse process reconstructing data from Gaussian noise via iterative denoising steps.}

We resort to the most popular text-to-image latent diffusion architecture \cite{rombach2022high} for leveraging its strong generalization power and adapting it to generate texture-consistent detail-rich normal maps. Some previous studies \cite{liu2024syncdreamer,shi2023zero123++,long2024wonder3d,wu2024unique3d} resort to joint multi-view diffusion to mitigate inconsistencies across views, yet in our implementation we choose the single-view working mode based on the following considerations: 
\begin{itemize}[leftmargin=23.5pt, topsep=1.0pt]
    \item[1)] Our targeted task of normal boosting estimates the detail patterns reflected by texture appearances based on given coarse normal maps, making the actual learning process more ``predictive'' than purely ``generative''. Moreover, there is no need to infer invisible contents and generate novel views from limited visual observations. Therefore, inter-view inconsistency can be expected to be controlled within reasonable ranges.
    \item[2)] The joint multi-view learning architecture requires fixed view configuration pre-determined during training. This setup restricts the flexibility in adjusting the number and positions of viewpoints if users have personalized needs (which turn to be quite common in practice) to designate specific surface areas for detail sculpting.
\end{itemize}

\vspace{0.5em}
\noindent\textbf{Straightforward Solution.} To achieve this goal, one straightforward diffusion-based solution is to denoise the noisy latent representation of the target normal map $\mathcal{H}$ while incorporating the color image $\mathcal{I}$ into the denoising U-Net. Thus, the forward process can be formulated as:
\begin{equation} \label{eqn:standard-forward}
    \widetilde{\mathcal{H}}_t = \sqrt{\bar{\alpha}_t} \cdot \widetilde{\mathcal{H}} + \sqrt{1 - \bar{\alpha}_t} \cdot \epsilon, \quad t = \{1,...,T\},
\end{equation}
\noindent where the noise schedule term $\bar{\alpha}_t$ controls the transformations to the standard Gaussian distribution $\epsilon$, $T$ is the number of time steps, $\widetilde{\mathcal{H}}$ denotes the latent representation of $\mathcal{H}$ after VAE's encoder, and $\widetilde{\mathcal{H}}_t$ is the noisy feature map. 

Given the denoising network $\mu_\theta$, the optimization objective can be formulated as:
\begin{equation} \label{eqn:standard-inverse}
    \mathcal{L}_\theta = \mathbb{E}_{\epsilon, \widetilde{\mathcal{I}}, \mathcal{H}, \mathbf{c}, t} \big[\|   \epsilon - \mu_\theta(\widetilde{\mathcal{I}}, \widetilde{\mathcal{H}}_t, \mathbf{c}, t) \|^2_2\big],
\end{equation}
\noindent where $\mathbf{c}$ is the vectorized embedding originally extracted from textual condition, and the visual cues can be injected by means of concatenating $\widetilde{\mathcal{H}}_t$ with $\widetilde{\mathcal{I}}$, the encoded 2D latent feature map of the color image $\mathcal{I}$.

\vspace{0.5em}
\noindent\textbf{Deterministic Diffusion.} In fact, such straightforward implementation as introduced above, together with various existing generative geometry modeling approaches \cite{long2024wonder3d,fu2024geowizard,wu2024unique3d}, can be regarded as cross-domain (RGB2Normal) image-to-image translation, while the special task of geometry super-resolution motivates us to refine the rendered coarse normal map $\mathcal{N}$ to generate the desired high-precision normal map $\mathcal{H}$.

To achieve stable and accurate normal detail generation, we introduce a deterministic diffusion mechanism by shifting the source and target domains, modeling a sequence of interpolations between coarse and target normal maps. Intuitively, this learning process can also be understood as multi-step image-to-image translations. In this way, the corresponding forward process can be formulated as:
\begin{equation}
    \widetilde{\mathcal{H}}_t = \sqrt{\bar{\alpha}_t} \cdot \widetilde{\mathcal{H}} + \sqrt{1 - \bar{\alpha}_t} \cdot \widetilde{\mathcal{N}}, \quad t = \{1,...,T\},
\end{equation}
\noindent where we directly replace the Gaussian noise $\epsilon$ (as in Eq.~(\ref{eqn:standard-forward})) with the latent representation $\widetilde{\mathcal{N}}$ of the rendered normal map $\mathcal{N}$ embedded from the VAE's encoder.

Eliminating the inherent stochastic nature of diffusion naturally ensures deterministic outputs and avoids the fluctuations of the generated normal maps when inferenced from multiple views, implicitly enhancing inter-view consistency and stabilizing the subsequent multi-view inverse rendering process.

\begin{figure*}[t!]
	\centering
	\includegraphics[width=0.8\linewidth]{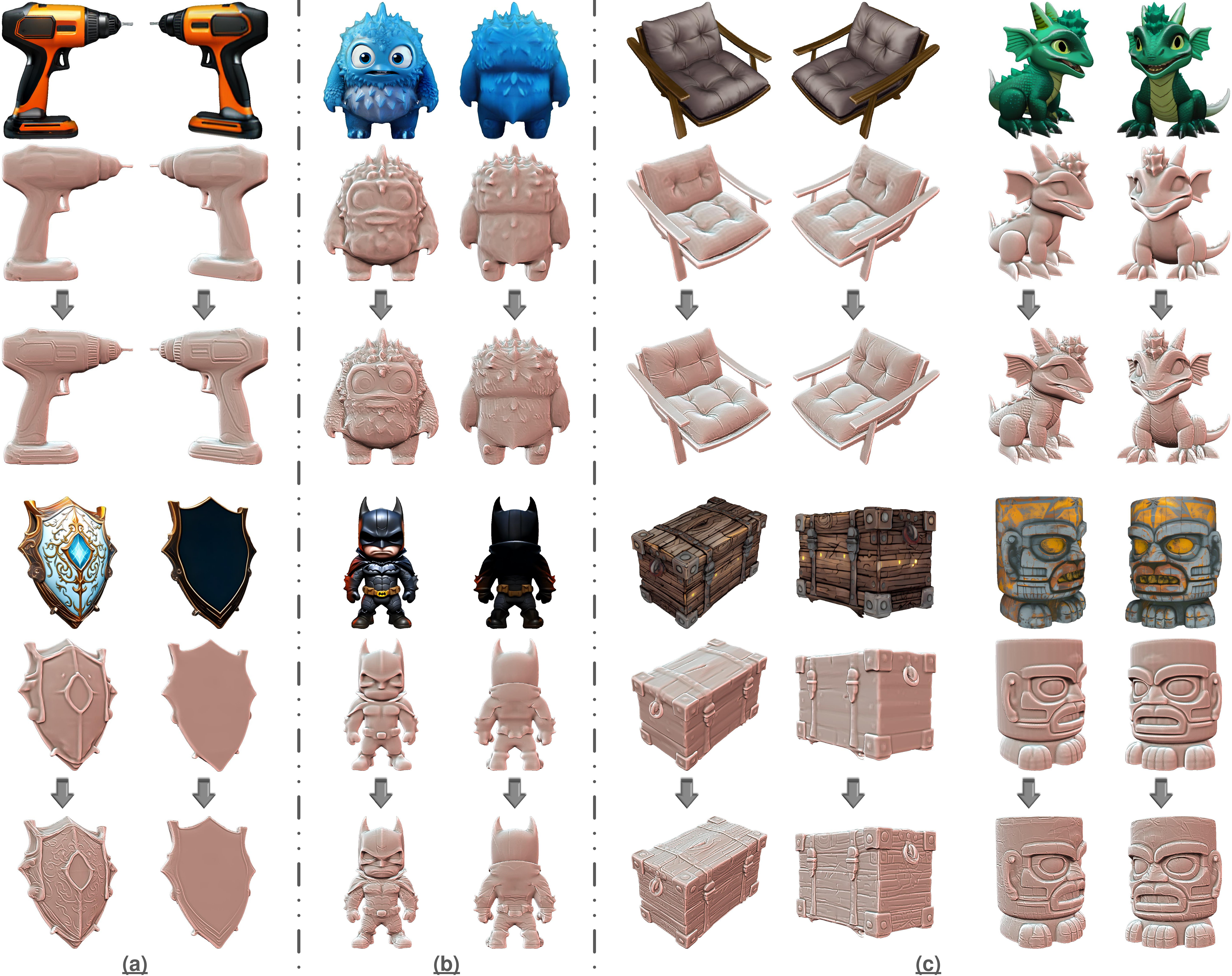}
	\caption{Illustration of applying our proposed SuperCarver for further 3D surface detail generation on textured meshes generated from (a) InstantMesh~\cite{xu2024instantmesh}, (b) Unique3D~\cite{wu2024unique3d}, and (c) CLAY~\cite{zhang2024clay}.}
	\label{fig:boosting-results}
	\vspace{-0.4cm}
\end{figure*}

\vspace{0.5em}
\noindent\textbf{Prior Extractions.} To enhance the accuracy and robustness of normal diffusion, we further extract semantic and geometric priors serving as the conditioning signals.

For the rendered color image $\mathcal{I}$, we extract its DINO~\cite{oquab2024dinov} feature $\widetilde{\mathcal{Q}}$ and also obtain its latent representation $\widetilde{\mathcal{I}}$ through VAE's encoder. The spatial resolution of $\widetilde{\mathcal{Q}}$ is pre-upscaled by several auxiliary convolutional layers. Besides, we employ CLIP's image encoder to extract the vectorized embedding $\tilde{\mathbf{c}}$ for cross-attention within U-Net's transformer blocks. For the rendered depth map $\mathcal{D}$, we utilize its latent representation $\widetilde{\mathcal{D}}$ as the geometric prior. Overall, we concatenate $\widetilde{\mathcal{Q}}$, $\widetilde{\mathcal{I}}$, and $\widetilde{\mathcal{D}}$ to deduce the resulting conditioning signal $\widetilde{\mathcal{S}}$.

Thus, we can achieve prior-guided diffusion by concatenating $\widetilde{\mathcal{S}}$ with the noisy latent representation $\widetilde{\mathcal{H}}_t$. The corresponding optimization objective can be formulated as:
\begin{equation}
    \mathcal{L}_\theta = \mathbb{E}_{\widetilde{\mathcal{N}}, \widetilde{\mathcal{S}}, \mathcal{H}, \tilde{\mathbf{c}}, t} \big[\|   \widetilde{\mathcal{N}} - \mu_\theta(\widetilde{\mathcal{S}}, \widetilde{\mathcal{H}}_t, \tilde{\mathbf{c}}, t)   \|^2_2\big].
\end{equation}

Empirically, we introduce another parameterization form of v-prediction~\cite{salimans2022progressive} for training the denoising U-Net, whose input channels and cross-attention dimensions are adapted to match with our prior-conditioned latent representations and the CLIP-encoded image embeddings.

\subsection{Noise-Resistant Inverse Normal Rendering} \label{sec:method:inverse-rendering}

\begin{figure*}[t!]
	\centering
	\includegraphics[width=0.8\linewidth]{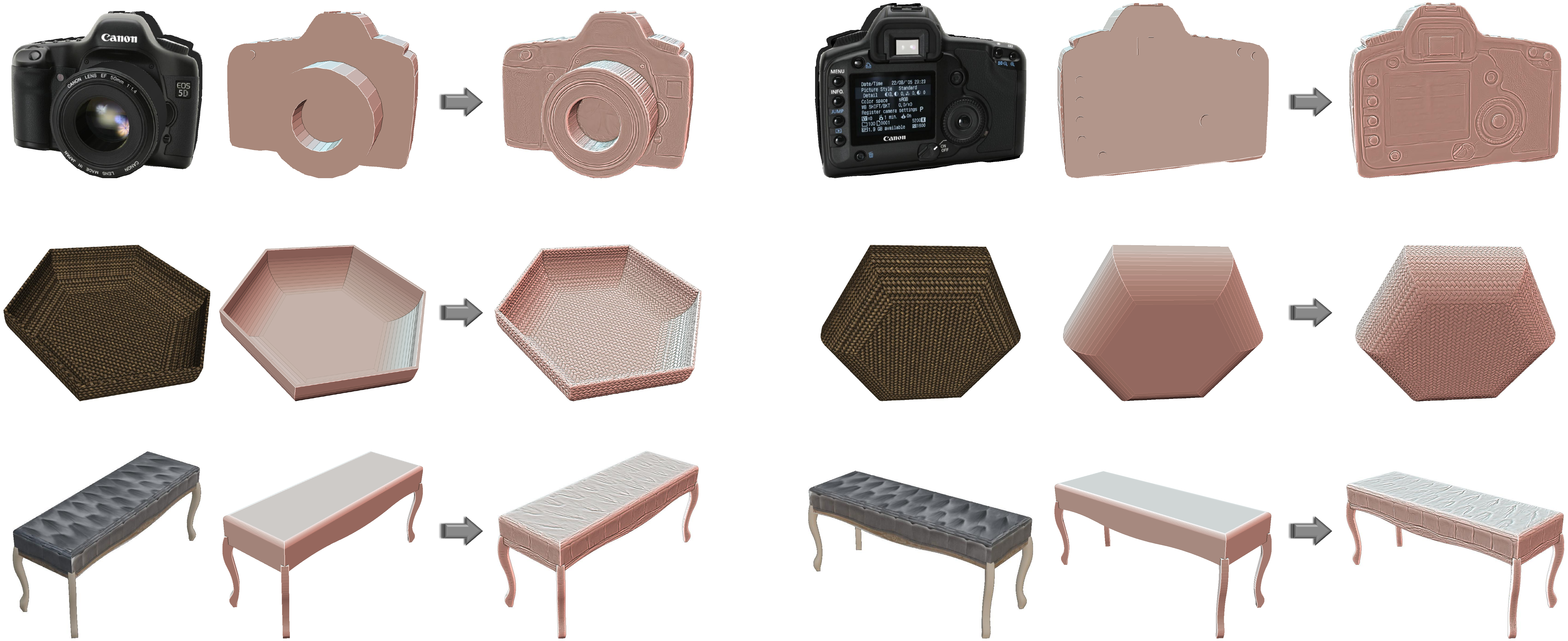}
	\caption{\revision{Results of applying our proposed SuperCarver on textured CAD mesh models collected from ShapeNet~\cite{chang2015shapenet}.}}
	\label{fig:cad_results}
	\vspace{-0.5cm}
\end{figure*}

\revision{By mathematically inverting the process of image formation, inverse rendering aims at estimating the underlying physical attributes of the target 3D scene, including geometry, textures,  materials, and lighting,  from 2D image observations.}

When finishing the training of our normal diffusion model, we can render multi-view 2D images from different viewpoints $\{ \mathbf{v}_k \}_{k=1}^K$, and perform inference to correspondingly predict a collection of detail-rich normal maps $\{ \mathcal{H}_k \in \mathbb{R}^{H \times W \times 3} \}_{k=1}^K$ to facilitate multi-view inverse rendering.

To transfer the underlying 3D structural patterns implied by the predicted 2D normal maps to the input mesh surface, the most common practice is to perform gradient-based mesh refinement through inverse rendering. However, it is observed that updating the mesh geometry from potentially imperfect normal map predictions is prone to causing obvious surface degradations, even with adaptations of various existing state-of-the-art (either explicit \cite{palfinger2022continuous} or implicit \cite{shen2023flexible}) mesh optimization approaches. Hence, we are strongly motivated to develop a highly noise-resistant scheme better suited to our setting of supplementing new details to a given coarse surface in a non-destructive manner.

\begin{figure}[t!]
	\centering
	\includegraphics[width=0.98\linewidth]{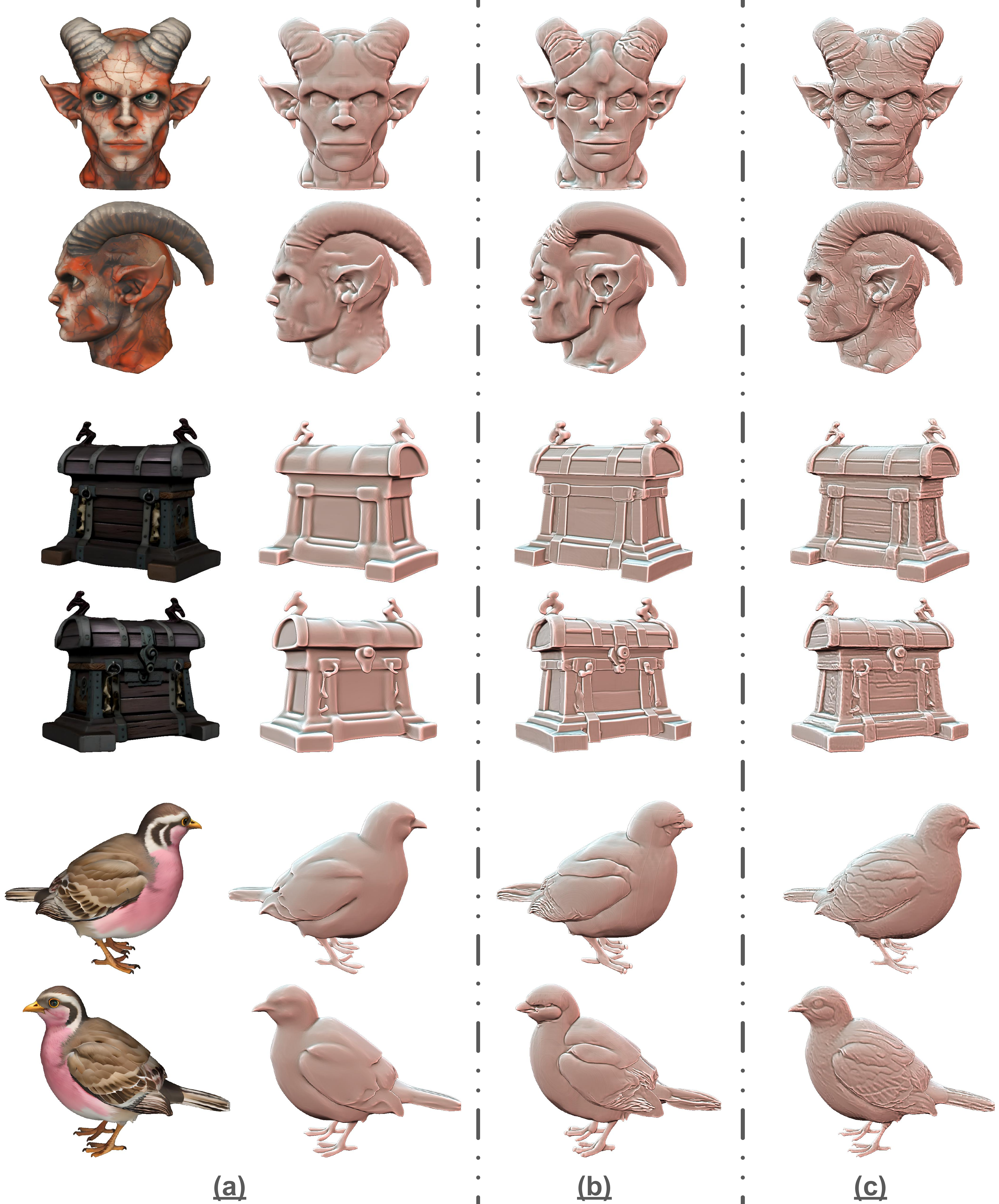}
	\caption{\revision{Comparison of geometry enhancement results produced by (b) DetailGen3D~\cite{deng2024detailgen3d} and (c) SuperCarver. The left column presents the (a) input textured meshes.}}
	\label{fig:compare_with_detailgen3d}
	\vspace{-1.0cm}
\end{figure}

We introduce a simple yet effective distance field deformation approach, which works in a \textit{``value-fixing, grid-moving''} manner. Specifically, assuming that the original mesh $\mathcal{M} = (\mathcal{V}, \mathcal{F})$ has already been normalized into the unit sphere, we start by constructing uniform 3D grid points $\mathcal{G} \in \mathbb{R}^{r \times r \times r \times 3}$ within $[-1, 1]^3$ and computing per-grid signed distance values $\mathcal{X} \in \mathbb{R}^{r \times r \times r}$. Different from common ways of implementing deformable distance field representations via only updating $\mathcal{X}$ or jointly updating both $\mathcal{G}$ and $\mathcal{X}$, we choose a subtle strategy that fixes the distance values $\mathcal{X}$ while separately learning zero-initialized offsets $\mathcal{O} \in \mathbb{R}^{r \times r \times r \times 3}$ to move initial grid positions $\mathcal{G}$. In comparison, standard distance field refinement requires higher resolutions of fixed uniform grids; otherwise, detailed structures cannot be well captured even with accurate distance values. Additionally, fixing pre-computed distance values hinders the optimization process from drastically changing the original mesh structure even with obviously erroneous normal prediction in certain areas.

To achieve this, we employ a differentiable implementation of the classic iso-surface extraction algorithm of DMC~\cite{nielson2004dual}, which we denote as $\Psi(\cdot; \cdot)$, for producing the updated mesh $\mathcal{M}^u = (\mathcal{V}^u, \mathcal{F}^u)$. Then, a differentiable rasterizer $\mathcal{R}(\cdot; \cdot)$ is integrated to render the correspondingly updated normal maps $\{ \mathcal{H}_k^u \in \mathbb{R}^{H \times W \times 3} \}_{k=1}^K$ with respect to each viewpoint $\mathbf{v}_k$. The whole processing procedure can be formulated as:
\begin{equation}
    \mathcal{G}^u = \mathcal{G} + \tau \cdot \sigma(\mathcal{O}),
\end{equation}
\begin{equation}
    \mathcal{H}_k^u = \mathcal{R}( \Psi(\mathcal{G}^u; \mathcal{X}) ; \mathbf{v}_k ),
\end{equation}
\noindent where we bound the range of grid offset values by applying the hyperbolic tangent (Tanh) function $\sigma(\cdot)$ multiplied by a positive scalar $\tau$, then the deformed grid points $\mathcal{G}^u$ and fixed distance values $\mathcal{X}$ are used for the subsequent differentiable normal map rendering.

\begin{figure*}[t!]
	\centering
	\includegraphics[width=0.8\linewidth]{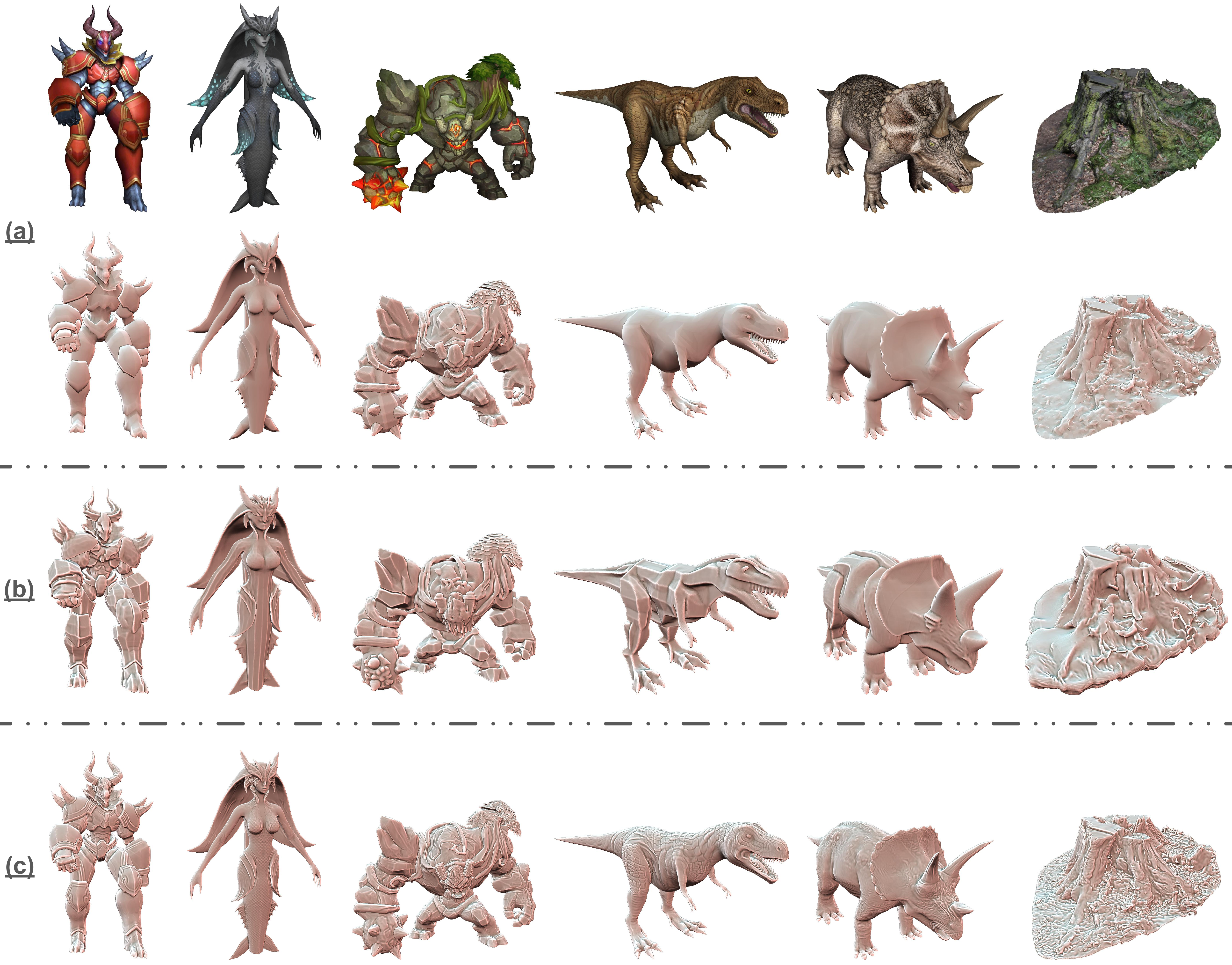}
	\vspace{-0.1cm}
	\caption{\revision{Comparison of geometry enhancement results produced by (b) DetailGen3D~\cite{deng2024detailgen3d} and (c) SuperCarver when evaluating more diverse distributions (\textit{e.g., humanoids, non-rigid, organic, scene scan}) of the (a) input textured meshes.}}
	\label{fig:richer_cases}
	\vspace{-0.5cm}
\end{figure*}

We iteratively optimize the grid offsets $\mathcal{O}$ by minimizing the following objective function:
\begin{equation}
    \min_{\mathcal{O}} \sum\nolimits_{k=1}^K \left\| \mathcal{H}_k^u - \mathcal{H}_k \right\|_2^2 + \omega_s \cdot \mathcal{L}_s + \omega_n \cdot \mathcal{L}_n,
\end{equation}
\noindent where the first term aligns all pairs of the diffusion-predicted and mesh-rendered normal maps, the last two terms denote mesh vertex and normal regularizations weighted by factors $\omega_s$ and $\omega_n$. For each vertex $\mathbf{p}_i$ in the updated mesh and its neighboring vertices $\Omega_i = \{ \mathbf{p}_{i,j} \}$, the Laplacian smoothing is computed by:
\begin{equation}
    \mathcal{L}_s = \sum\nolimits_{\mathbf{p}_i \in \mathcal{V}^u} \sum\nolimits_{\mathbf{p}_{i,j} \in \Omega_i} (\mathbf{p}_i - \mathbf{p}_{i,j}) / \vert \Omega_i \vert,
\end{equation}
\noindent where $\vert \Omega_i \vert$ is the number of neighbors. Moreover, for each pair of neighboring normal vectors $\mathbf{n}_i$ and $\mathbf{n}_i^{\prime}$, we constrain local orientation consistency by:
\begin{equation}
    \mathcal{L}_n = \sum\nolimits_i  1 - (\mathbf{n}_i \cdot  \mathbf{n}_i^{\prime}) / (\left\| \mathbf{n}_i \right\|  \cdot  \left\| \mathbf{n}_i^{\prime} \right\|).
\end{equation}

After iterative optimization, we moderately apply the Taubin smoothing \cite{taubin1995signal} as a post-processing step to mitigate meaningless surface wrinkles.

\section{Experiments} \label{sec:experiments}

This section begins by introducing our main technical implementations in terms of data preparation and framework hyperparameters. \revision{For performance evaluations, we qualitatively illustrate the boosting effects of applying the proposed SuperCarver to a collection of diverse real-world 3D assets as well as synthesized mesh models from different data distributions. Comprehensive comparisons are conducted to show the advantages of our framework against other relevant approaches.} To validate the superiority of our core technical components, we make comparisons with state-of-the-art normal prediction and mesh optimization approaches. We further conduct a series of ablation studies and analyses. In the end, we also discuss our current limitations and potential future explorations.

\vspace{0.5em}
\noindent\textbf{Training Dataset Preparation.} We used the publicly available Objaverse~\cite{deitke2023objaverse} dataset comprising 800K+ 3D assets to produce our training data. Through a series of meticulously designed filtering protocols and cleaning-up procedures, we ultimately acquired around 80K high-quality mesh models.

We used BlenderProc~\cite{denninger2023} for the rendering of color images, depth maps, and normal maps based on perspective projection. For each training mesh, we randomly sampled 16 viewpoints, with azimuth and elevation angles respectively varying within [0$^\circ$, 360$^\circ$] and [-30$^\circ$, 30$^\circ$].

To create the paired low-poly and high-poly normal renderings, we performed geometry simplification on each high-poly training mesh model based on manifold reconstruction \cite{huang2018robust}, and further removed detailed structures using \cite{taubin1995signal}. We further manually annotated a smaller portion of unqualified training samples, which were used for training a separate neural model to purify the whole training set.

\begin{figure}[t!]
	\centering
	\includegraphics[width=0.98\linewidth]{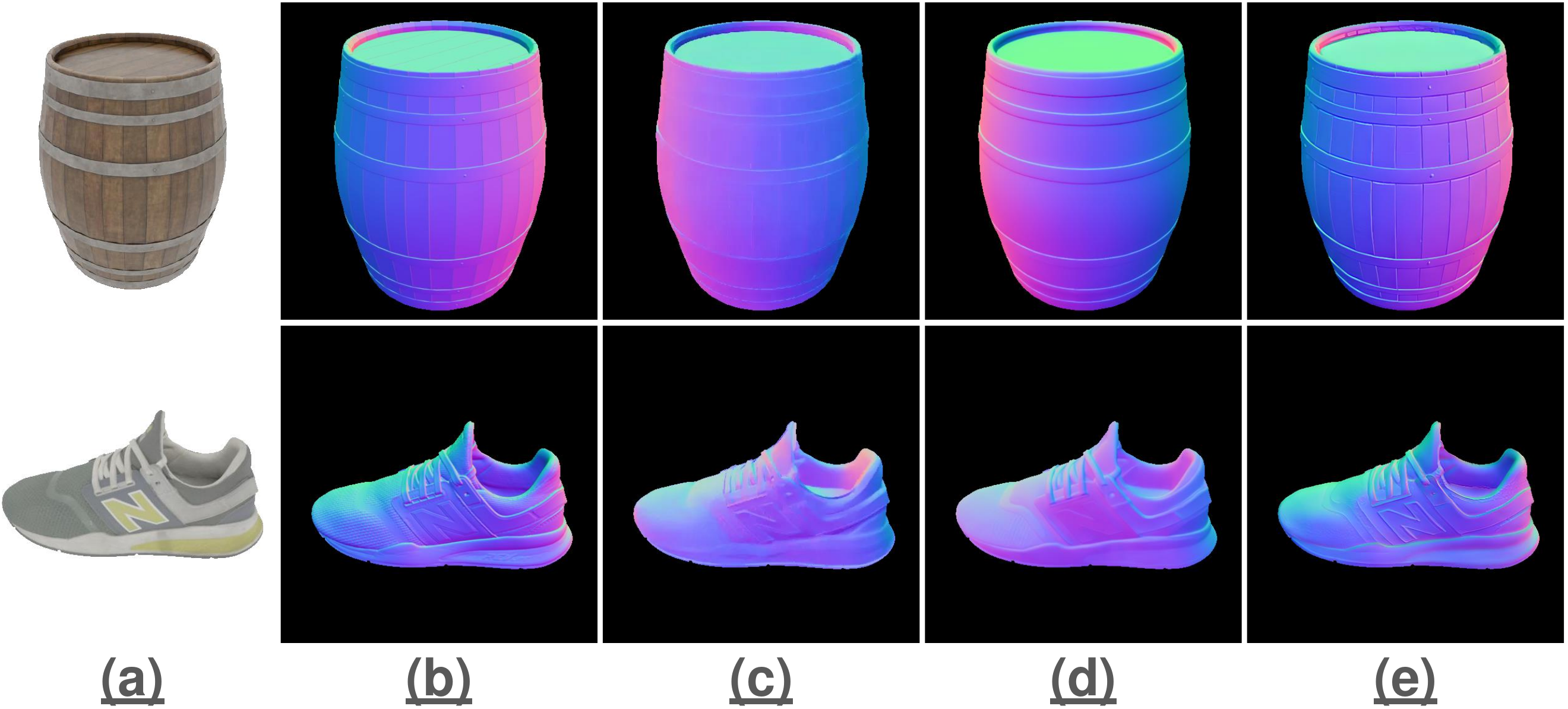}
	\caption{Results of normal map predictions generated from (c) GeoWizard~\cite{fu2024geowizard}, (d) StableNormal~\cite{ye2024stablenormal}, and (e) Ours, where (a) and (b) show the corresponding color images and ground-truth high-fidelity normal maps.}
	\label{fig:compare_normal_results}
	\vspace{-0.5cm}
\end{figure}

\begin{figure}[t!]
	\centering
	\includegraphics[width=0.98\linewidth]{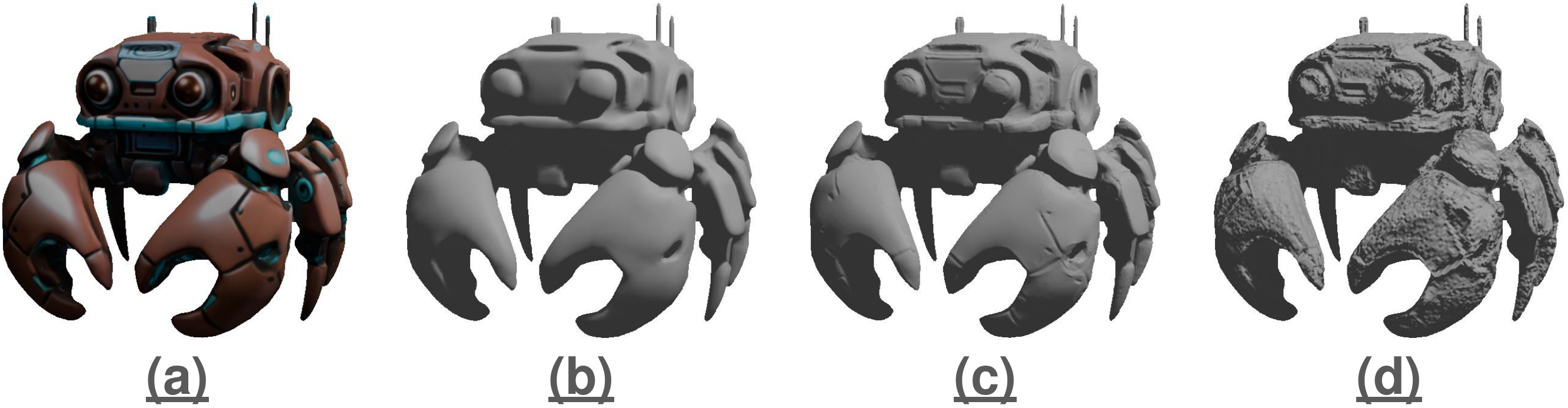}
	\caption{Rendering effects of (c) our optimized mesh and (d) attaching the bump map obtained from PBRBoost~\cite{palfinger2022continuous}, given the input textured coarse mesh in (a) and (b).}
    \vspace{-0.5cm}
	\label{fig:compare_pbrboost}
\end{figure}

\vspace{0.5em}
\noindent\textbf{Implementation Details.} We employed the popular diffusion architecture of Stable~Diffusion~V2.1 \cite{rombach2022high} and performed fine-tuning on 16 NVIDIA Tesla V100 GPUs for 40K steps with a total batch size of 128. We applied the AdamW~\cite{loshchilov2018decoupled} optimizer with cosine annealing, reducing the learning rate from 5e-5 to 1e-5. For both training and inference, the image resolution is uniformly configured to 512$\times$512.

For iterative triangular mesh optimization, we adopted Nvdiffrast~\cite{laine2020diffrast} for efficient differentiable rendering from 12 camera poses (8 evenly-spaced horizontal views with azimuth angles from 0$^\circ$ to 360$^\circ$ and 4 downward-looking (elevated by 30$^\circ$) views with azimuth angles [45$^\circ$, 135$^\circ$, 225$^\circ$, 315$^\circ$]). Camera intrinsics were kept the same as our configurations in BlenderProc. The hyperparameters were empirically tuned as $r=512$, $\tau=0.5$, $\omega_s=0.25$, $\omega_n=0.01$ for 200 iterations.

When deployed on NVIDIA L20 GPU (mid-range product for AI inference), the average time consumption for the whole geometry super-resolution workflow is around 1$\sim$2 minutes, which is thought to be adequately efficient.

To facilitate quantitative evaluations, we manually selected 50 detail-rich textured meshes as our ground-truths from the widely-used Objaverse~\cite{deitke2023objaverse} and ABO~\cite{collins2022abo} shape repositories. We applied surface simplification and smoothing to produce geometrically-coarse meshes as the input samples.

\subsection{Performance Evaluation}

\begin{figure}[t!]
    \centering
    \includegraphics[width=1.0\linewidth]{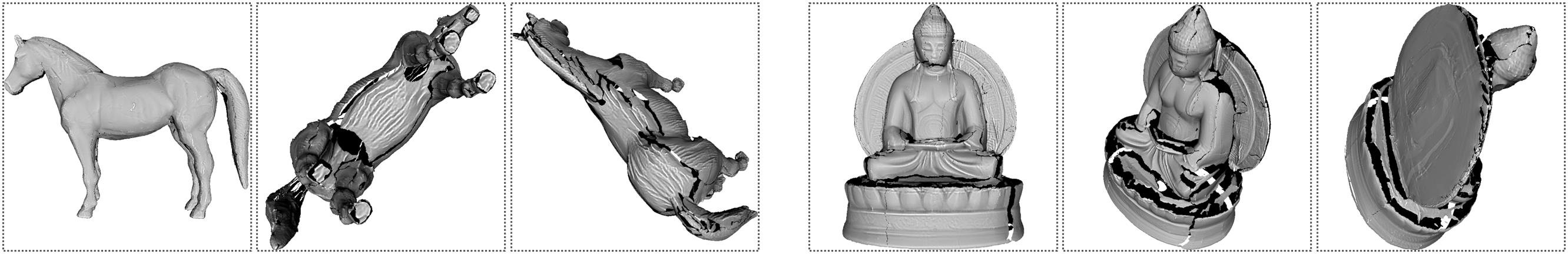}
    \caption{Typical surface damages produced by the optimization process of FlexiCubes~\cite{shen2023flexible}.}
    \label{fig:flexicubes_damage}
    \vspace{-0.3cm}
\end{figure}

\begin{figure}[t!]
    \centering
    \includegraphics[width=0.98\linewidth]{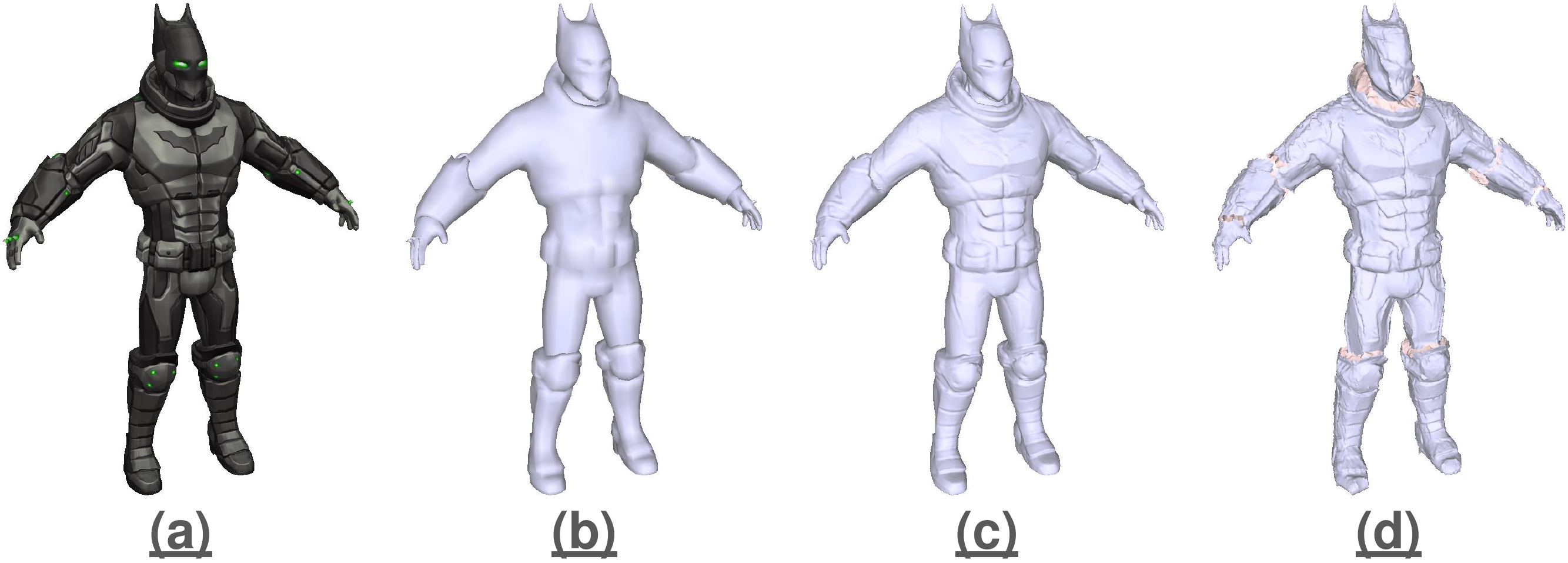}
    \caption{Comparison of mesh optimization effects, where (a) and (b) denote the input textured mesh, (c) and (d) present the results of our distance field deformation scheme and CR~\cite{palfinger2022continuous}.}
    \label{fig:compare_mesh_opt}
    \vspace{-0.5cm}
\end{figure}

We presented the qualitative visualizations for the outputs of applying SuperCarver to diverse geometrically-coarse textured meshes. As illustrated in Figure~\ref{fig:more-results}, our approach is capable of supplementing realistic and expressive surface details under a variety of testing cases.

In addition to real-world 3D assets, we further experimented with AI-generated mesh assets by applying SuperCarver to the textured mesh models generated from advanced 3D generation frameworks. We adopted InstantMesh~\cite{xu2024instantmesh}, Unique3D~\cite{wu2024unique3d}, and CLAY~\cite{zhang2024clay}, covering representative generation paradigms of feed-forward large reconstruction models, multi-view diffusion models, and native 3D latent diffusion models. As compared in Figure~\ref{fig:boosting-results}, various input meshes are effectively boosted by our approach to produce realistic and expressive 3D surface details well aligned with texture appearances. \revision{Furthermore, we evaluated textured CAD meshes in the classic ShapeNet~\cite{chang2015shapenet} repository (note that typical textureless CAD meshes such as those in the classic ABC~\cite{koch2019abc} dataset are not the type of input our approach is designed for). As illustrated in Figure~\ref{fig:cad_results}, our approach successfully produces high-fidelity detail patterns.}

\revision{In particular, we made comprehensive comparisons with the more recent competing approach of DetailGen3D~\cite{deng2024detailgen3d} targeted at 3D geometry enhancement. As depicted in Figures~\ref{fig:compare_with_detailgen3d} and \ref{fig:richer_cases}, DetailGen3D cannot sufficiently preserve the original structure of input meshes and the generated geometric patterns are much less expressive, because its image guidance effects are rather weak and ambiguous. We further conducted quantitative evaluations in Table~\ref{tab:compare_with_detailgen3d}, where our approach consistently outperforms DetailGen3D across all the involved 2D (MAE) and 3D (CD, F-Score) metrics. On average, DetailGen3D requires around $1$ minute for geometry enhancement, while our framework takes around $1.5$ minutes. Although our framework design involves iterative optimization, its actual running efficiency is basically satisfactory and comparable to DetailGen3D.}

\begin{table*}[t!]
	\centering	
	\renewcommand\arraystretch{1.25}
	\setlength{\tabcolsep}{30.0pt}
	\caption{\revision{Quantitative comparison of geometry enhancement results produced by DetailGen3D~\cite{deng2024detailgen3d} and SuperCarver, including the 2D metric of MAE, the 3D metrics of Chamfer Distance (CD) and F-Score (following the implementation adopted in \cite{wang2018pixel2mesh}), as well as the overall running efficiency (evaluated on the same device environment with an NVIDIA L20 GPU).}}
	\vspace{-0.15cm}
	\begin{tabular}{ c | c | c | c | c}
		\toprule[1.0pt]
		Method & MAE $\downarrow$ & CD ($\times 10^{-4}$) $\downarrow$ & F-Score $\uparrow$ & Time Cost \\
		\hline
		DetailGen3D~\cite{deng2024detailgen3d} & 9.11$^\circ$ & 1.84 & 57.31\% & $\sim\,$\textbf{58s} \\
		Ours & \textbf{4.22$^\circ$} & \textbf{0.41} & \textbf{86.18\%} & $\sim\,$95s \\
		\bottomrule[1.0pt]
	\end{tabular}
	\label{tab:compare_with_detailgen3d}
	\vspace{-0.15cm}
\end{table*}

\vspace{0.5em}
\noindent\textbf{Comparison on Normal Prediction.} To evaluate the effectiveness of our specialized normal diffusion architecture, we made comparisons with two advanced normal diffusion frameworks, including GeoWizard~\cite{fu2024geowizard} and StableNormal~\cite{ye2024stablenormal}.

As compared in Figure~\ref{fig:compare_normal_results}, our normal map predictions show better details and avoid obviously erroneous areas. Besides, we further deduced the metric of mean angular error (MAE) between ground-truth and predicted normal maps, where we masked  invalid backgrounds for the output images of GeoWizard and StableNormal. As compared in Table~\ref{tab:quant_comparison_normal_pred}, thanks to our specialized diffusion architecture, our normal map predictions significantly outperform the competing approaches.

\begin{table}[t!]
    \centering	
    \renewcommand\arraystretch{1.25}
    \setlength{\tabcolsep}{10.9pt}
    \caption{Quantitative comparison of normal map prediction accuracy under the metric of mean angular error (MAE).}
    \vspace{-0.15cm}
    \begin{tabular}{ c | c | c | c }
        \toprule[1.0pt]
        Method & GeoWizard~\cite{fu2024geowizard} & StableNormal~\cite{ye2024stablenormal} & Ours \\ 
        \hline
        MAE $\downarrow$ & 17.59$^\circ$ & 14.33$^\circ$ & \textbf{9.05$^\circ$} \\ 
        \bottomrule[1.0pt]
    \end{tabular}
    \label{tab:quant_comparison_normal_pred}
\end{table}

\begin{table}[t!]
    \centering	
    \renewcommand\arraystretch{1.25}
    \setlength{\tabcolsep}{11.4pt}
    \caption{Normal map diffusion accuracy of different baseline schemes. ``\textit{Stoc.}'' denotes the straightforward implementation of stochastic diffusion, then ``\textit{w/o Sem.}'' and ``\textit{w/o Geo.}'' respectively represent removing the high-level semantics and low-level depth priors.}
    \vspace{-0.15cm}
    \begin{tabular}{ c | c | c | c | c }
        \toprule[1.0pt]
        Method & Original & \textit{Stoc.} & \textit{w/o Sem.} & \textit{w/o Geo.} \\
        \hline
        MAE $\downarrow$ & \textbf{9.05$^\circ$} & 21.64$^\circ$ & 10.49$^\circ$ & 9.81$^\circ$ \\
        \bottomrule[1.0pt]
    \end{tabular}
    \label{tab:ablation_normal_diffusion}
\end{table}

\begin{table}[t!]
    \centering	
    \renewcommand\arraystretch{1.25}
    \setlength{\tabcolsep}{14.5pt}
    \caption{Quantitative comparison of our different baseline design choices for achieving distance field deformation. ``\textit{Only Grids}'' represents the original design, ``\textit{Only Dist.}'' and ``\textit{Joint G\&D}'' are the corresponding two variants.}
    \vspace{-0.15cm}
    \begin{tabular}{ c | c | c | c }
        \toprule[1.0pt]
        Method & \textit{Only Grids} & \textit{Only Dist.} & \textit{Joint G\&D}  \\
        \hline
        MAE $\downarrow$ & \textbf{4.22$^\circ$} & 4.75$^\circ$ & 5.47$^\circ$ \\
        \bottomrule[1.0pt]
    \end{tabular}
    \label{tab:ablation_sdf_opt}
\end{table}

\begin{table}[t!]
    \centering	
    \renewcommand\arraystretch{1.25}
    \setlength{\tabcolsep}{8.0pt}
    \caption{Quantitative evaluation of cross-view consistency under the metric of mean angular error (MAE).}
    \vspace{-0.15cm}
    \begin{tabular}{ c | c }
        \toprule[1.0pt]
        \textit{Overall Normal Diffusion Error} & \textit{Cross-View Normal Inconsistency}  \\
        \hline
        9.05$^\circ$ & 7.24$^\circ$ \\
        \bottomrule[1.0pt]
    \end{tabular}
    \label{tab:ablation_view_consistency}
    \vspace{-0.2cm}
\end{table}

In addition, we also included PBRBoost~\cite{wang2024boosting} (an advanced material generation framework), since one of its outputs is the 2D bump maps associated with the base meshes, enabling us to compare its rendering effects when attaching the bump maps as height information. As shown in Figure~\ref{fig:compare_pbrboost}, the actual fidelity of the enhanced geometric cues from PBRBoost is still far from satisfactory.

\vspace{0.5em}
\noindent\textbf{Comparison on Mesh Optimization.} To facilitate 3D surface geometry updating, many recent mesh generation frameworks \cite{lu2024direct2,choi2024ltm,li2024craftsman,wu2024unique3d} introduce the CR~\cite{palfinger2022continuous} algorithm with joint optimization and remeshing, which is regarded as the most relevant competitor. In fact, we had previously attempted to adapt FlexiCubes~\cite{shen2023flexible}, an advanced iso-surface representation, to achieve gradient-based geometry refinement starting from the initially given mesh surface. However, through extensive attempts, we empirically found that it easily leads to severe surface damages, as illustrated in Figure~\ref{fig:flexicubes_damage}. We reason that  this is attributable to its significantly increased deformation flexibility (which should have been its major strength) amplifies the impacts of erroneous normal pixels. As shown in Figure~\ref{fig:compare_mesh_opt}, for typical hard cases with less accurate normal predictions, CR tends to produce noisy and degraded surfaces.

\begin{figure}[t!]
	\centering
	\includegraphics[width=0.98\linewidth]{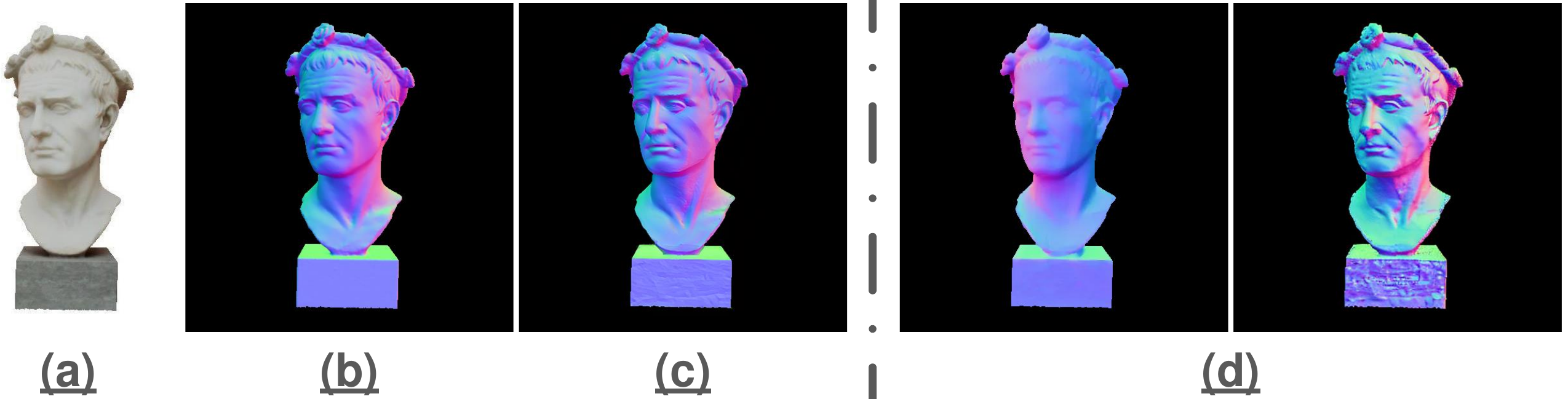}
	\caption{Illustration of prediction variation of stochastic baseline diffusion as displayed in (d). The color image, ground-truth normal map, and the output of our original design are presented in (a), (b), and (c).}
	\label{fig:randomness_of_baseline}
	\vspace{-0.4cm}
\end{figure}

\begin{figure}[t!]
	\centering
	\includegraphics[width=0.98\linewidth]{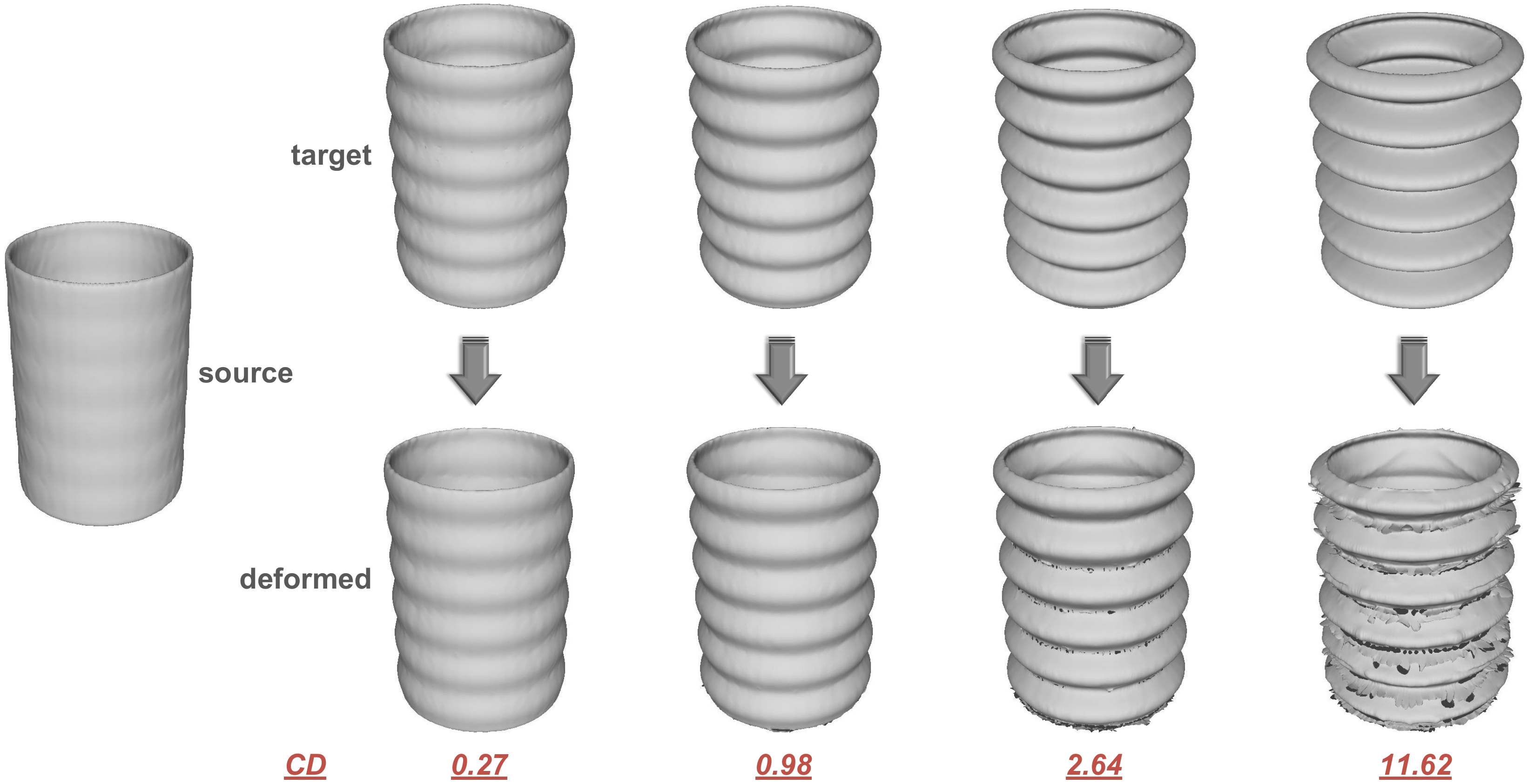}
	\caption{\revision{Robustness of our distance field deformation approach when deforming the same source mesh to target meshes with  increasingly larger deformation levels, with the corresponding Chamfer Distance (CD, $\times 10^{-4}$) measurement reported.}}
	\label{fig:large_deform_test}
	\vspace{-0.75cm}
\end{figure}

\subsection{Ablation Study}

To evaluate the effectiveness and necessity of the core components in our entire workflow, we conducted ablation study with different baseline schemes. We began with a straightforward baseline normal diffusion framework as formulated in Eq.~(\ref{eqn:standard-forward}) and Eq.~(\ref{eqn:standard-inverse}). Besides, since prior guidance is also a critical condition for generating robust and accurate normal maps, we can design another two variants via removing the semantic cues extracted from DINO features or the geometric cues extracted from depth maps.

As reported in Table~\ref{tab:ablation_normal_diffusion}, directly fine-tuning a standard diffusion model conditioned on color images lead to significantly degraded performance, with the MAE metric of 21.64$^\circ$. More importantly, as illustrated in Figure~\ref{fig:randomness_of_baseline}, the stochastic nature inevitably results in unstable predictions. As reported in the last two columns of Table~\ref{tab:ablation_normal_diffusion}, the joint injection of semantic and geometric priors are indeed beneficial. 

For inverse normal rendering, we experimented with another two distance field deformation alternatives, i.e., fixing grids while optimizing distances; jointly updating both grids and distances. To produce quantitative results, we computed the MAE metric of normal maps \textit{rendered} from optimized meshes (note that Table~\ref{tab:quant_comparison_normal_pred} and Table~\ref{tab:ablation_normal_diffusion} are targeted at \textit{diffused} normal maps). As evidenced in Table~\ref{tab:ablation_sdf_opt}, the two variants lead to sub-optimal performances, in that only optimizing distances usually causes weakened details while the joint optimization strategy easily suffers from surface distortion especially when the generated normal maps contain relatively larger errors. \revision{Besides, although our targeted geometry super-resolution task devotes to generating realistic geometric details that represent \textbf{\textit{mild}} deviation from the original surface, we still experimented with large deformation scenarios to explore the characteristics of our ``value-fixing, grid-moving'' strategy. As shown in Figure~\ref{fig:large_deform_test}, our scheme cannot robustly deal with significant mesh deformations, i.e., producing obvious surface degradations and outliers. Through empirical explorations, our proposed strategy maintains satisfactory performance for vertex offsets bounded by approximately \textbf{\textit{0.03}}.}

\begin{figure}[t!]
	\centering
	\includegraphics[width=0.98\linewidth]{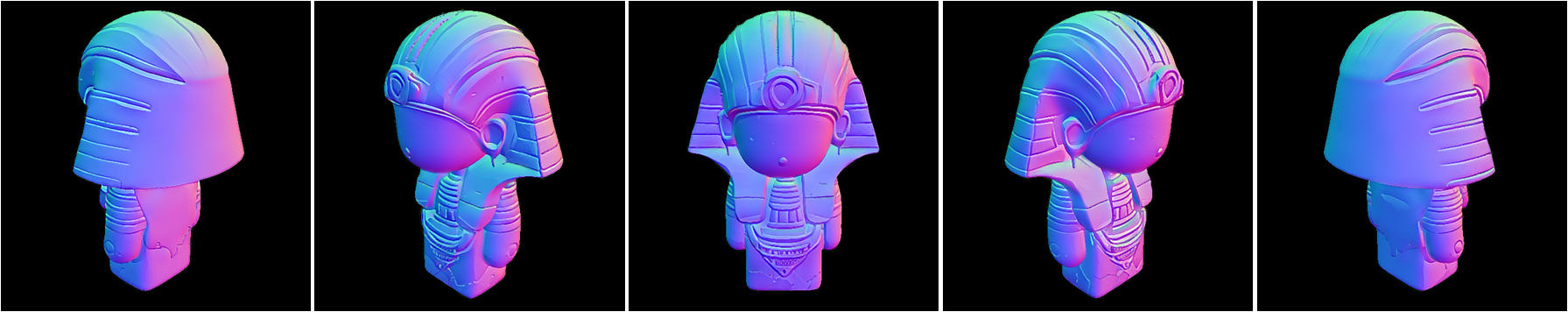}
	\caption{Illustration of our multi-view normal predictions.}
	\label{fig:our_view_consistency}
	\vspace{-0.2cm}
\end{figure}

Additionally, we quantitatively measured normal deviations across views in Table~\ref{tab:ablation_view_consistency}, in which the average MAE is 7.24$^\circ$. Comparing this metric with the overall normal prediction error with respect to ground-truth normal maps and the overall performances of other state-of-the-art competitors as reported in Table~\ref{tab:quant_comparison_normal_pred}, we can observe that our view consistency performance is satisfactory enough, as visualized in Figure~\ref{fig:our_view_consistency}.

\subsection{\revision{Limitations, Failure Cases, and Future Works}}

\begin{figure}[t!]
	\centering
	\includegraphics[width=0.85\linewidth]{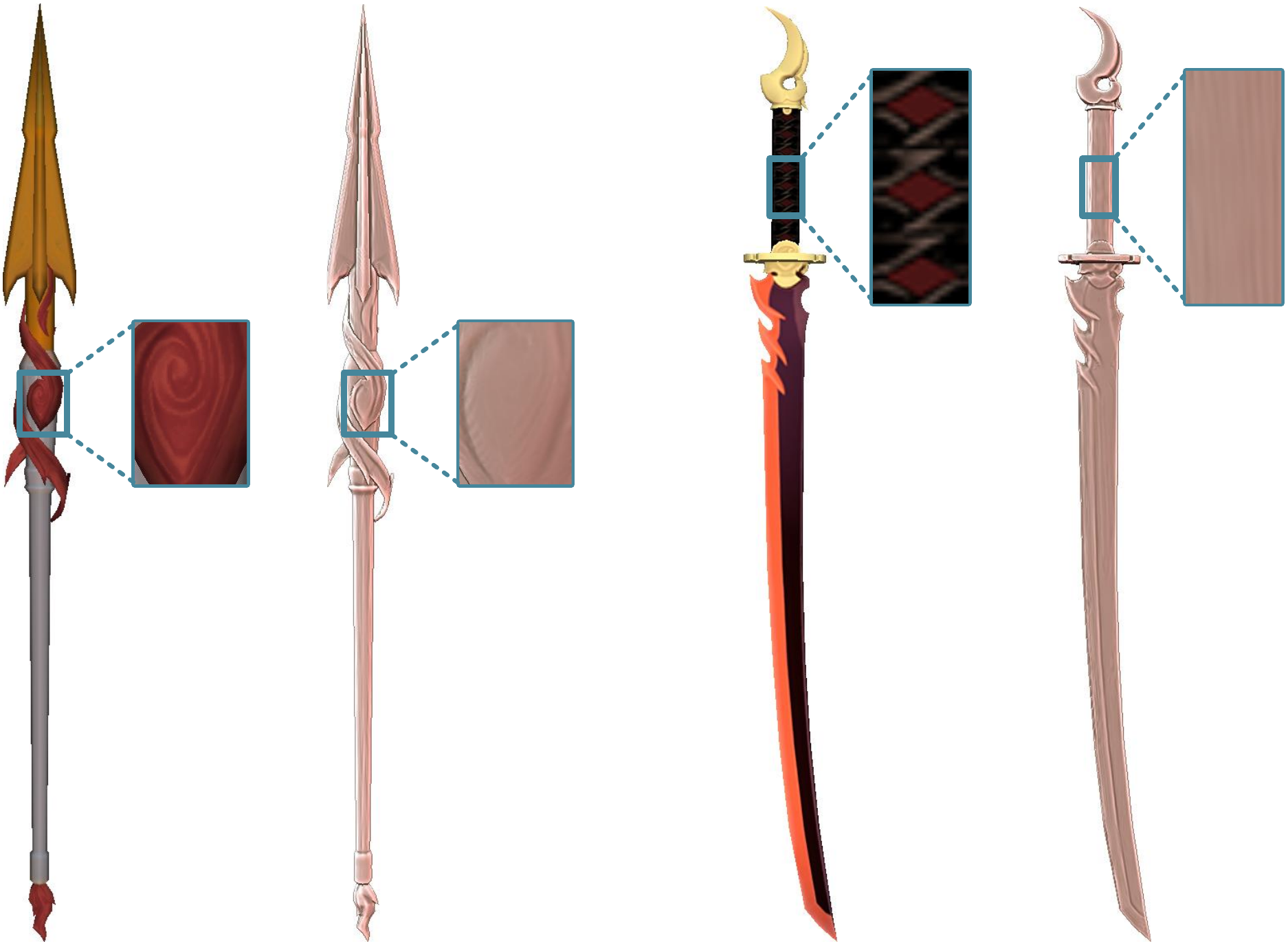}
	\caption{\revision{Results of applying SuperCarver on elongated objects.}}
	\label{fig:thin_cases}
	\vspace{-0.3cm}
\end{figure}

\begin{figure}[t!]
	\centering
	\includegraphics[width=1.0\linewidth]{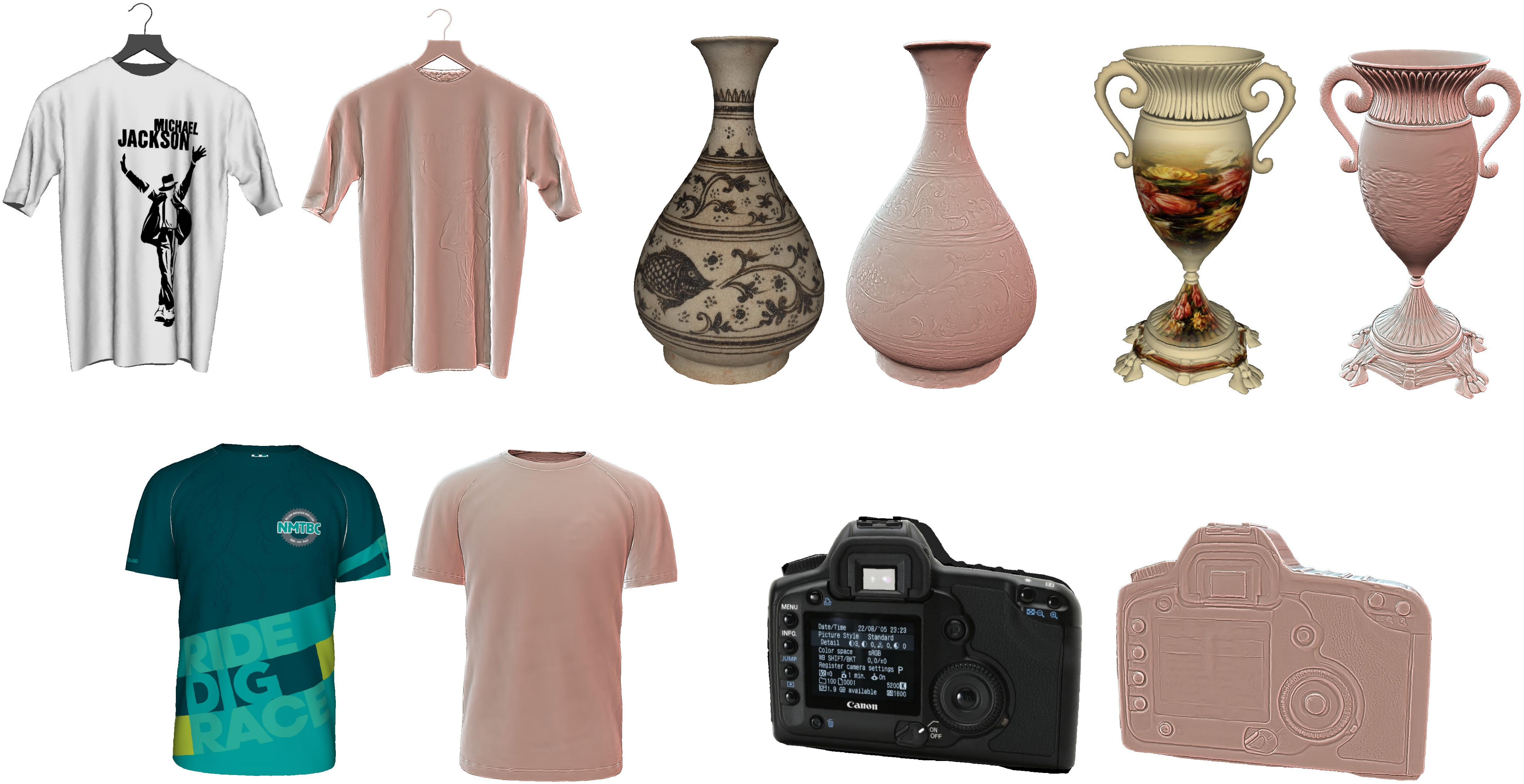}
	\caption{\revision{Illustration of potential texture overfitting caused by SuperCarver. The first row shows three failure cases with obvious texture overfitting. The second row shows two examples where texture overfitting issues are basically negligible.}}
	\label{fig:ours_texture_overfitting}
	\vspace{-0.3cm}
\end{figure}

\begin{figure}[t!]
	\centering
	\includegraphics[width=1.0\linewidth]{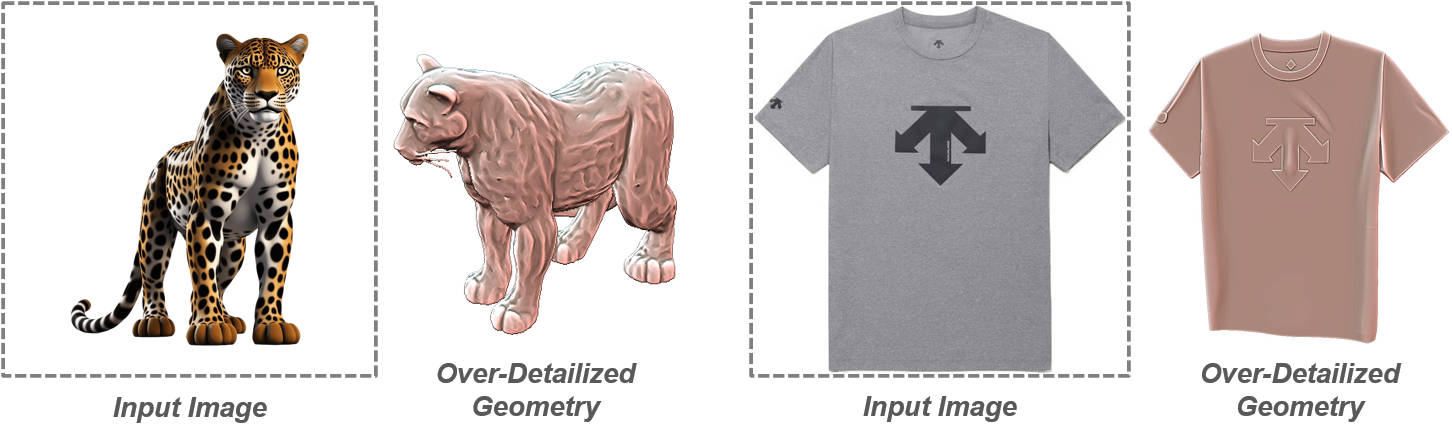}
	\caption{Illustration of over-detailization in Hunyuan3D~\cite{zhao2025hunyuan3d}.}
	\label{fig:hunyuan_texture_overfitting}
	\vspace{-0.5cm}
\end{figure}

\revision{As a novel learning-based solution for texture-consistent 3D geometry super-resolution, SuperCarver demonstrates satisfactory fidelity and applicability. Nevertheless, as a preliminary attempt in this direction, our current technical implementation still has several aspects of limitations for future explorations.}

\revision{With finite image resolution, our normal boosting component cannot sufficiently enhance the detail patterns of highly elongated objects (e.g., knives, swords), as shown in Figure~\ref{fig:thin_cases}. The reason is that the rendered areas are too small to provide adequate appearance information and visual cues. To overcome this issue, we might need to construct a localized processing pipeline where the original 3D mesh is rendered and detailized through adaptive sliding windows.}

\revision{Hindered by inconsistent data quality of the training meshes, there inevitably exist over-detailization phenomena. As shown in Figure~\ref{fig:ours_texture_overfitting}, our results show additional geometric details in the three cases of the first row. However, since our approach can implicitly learn  discrimination capability from the training data distribution, it succeeds in avoiding texture overfitting in the last two cases of the second row. Theoretically, identifying  geometry-irrelevant texture patterns is supposed to be unified within the overall learning process when training with large volumes of mesh data. Therefore, more stringent data selection protocols are required to ensure geometry-texture consistency in the training samples. In addition, in certain cases, the same texture semantically allows for multiple valid interpretations of 3D surface details. Although our deterministic framework significantly improves stability and multi-view inconsistency, it inherently sacrifices diversity to some extent when dealing with geometry-appearance ambiguity.}
 
\revision{Nevertheless, it is worth mentioning that texture overfitting remains unresolved in the current 3D generation community. Even the current state-of-the-art proprietary foundation model \cite{zhao2025hunyuan3d} suffers from the occurrence of the overly detailized 3D geometric structures as depicted in Figure~\ref{fig:hunyuan_texture_overfitting}.} 

\secondrevision{Particularly, it is worth noting that although the latest state-of-the-art industrial-grade 3D mesh generation products, such as Hunyuan3D 3.1, Tripo 3.1, and Meshy 6, have been capable of generating highly expressive geometries, it is still hard to avoid situations where fine surface details are insufficiently generated, especially for complex 3D models with rich texture patterns. As compared in Figure~\ref{fig:enhance_industrial_products}, applying our SuperCarver can further bring noticeable improvement in terms of detailed surface geometry fidelity.}

\revision{Additionally, the inherent properties of our adopted \textbf{\textit{signed}} distance field representation structure restrict our capabilities to handle open surfaces and/or highly complicated topologies. Hence, we need to explore differentiable iso-surface extractors built upon more generic implicit geometry representations such as appropriate adaptations of \textbf{\textit{unsigned}} distance fields.}

\revision{Finally, our second stage of mesh detail generation relies on iterative gradient-based optimization. One promising direction is to build feed-forward neural architectures to further improve the geometric quality and inference efficiency. Considering the normal detailization stage as a 2D branch and the subsequent mesh optimization stage as a 3D branch, we could attempt to re-design the 3D branch into a feed-forward architecture. For example, we could begin by extracting multi-view 2D features from both the rendered RGB images and the boosted normal maps, then back-projecting and merging pixel features to the 3D domain, forming a 3D feature volume. To achieve surface detailization, we could append a prediction head that consumes the grid features and directly regress the desired 3D grid offsets under the supervision of pre-computed ground-truth values.}

\begin{figure*}[t!]
	\centering
	\includegraphics[width=0.95\linewidth]{./figs/enhance_industrial_products.pdf}
	\caption{\secondrevision{Illustration of applying SuperCarver to the generated textured meshes of Hunyuan3D 3.1, Tripo 3.1, and Meshy 6.}}
	\label{fig:enhance_industrial_products}
	\vspace{-0.5cm}
\end{figure*}

\section{Conclusion} \label{sec:conclusion}

This paper presents a specialized geometry super-resolution framework for high-fidelity texture-consistent 3D surface detail generation. The proposed SuperCarver pipeline is normal-centric and composed of two sequential stages. The first stage generates detail-boosted 2D normal maps from multiple views via deterministic prior-guided normal diffusion. The second stage transfers the underlying geometric patterns of multi-view normal map predictions onto the original 3D mesh surface via noise-resistant inverse rendering driven by carefully designed distance field deformation. Extensive experiments demonstrate the effectiveness and applicability of SuperCarver across various asset types and texture styles.

Our approach serves as a powerful tool for high-poly surface sculpting and historical low-quality mesh geometry upgrading. We believe that our investigations would inspire more focused researches in the field of 3D geometry super-resolution, which is highly valuable yet rarely explored in the current community of 3D generative modeling.

\bibliographystyle{IEEEtran}
\bibliography{ref}

\vspace{-0.3cm}
\begin{IEEEbiography}[{\includegraphics[width=1in,height=1.25in,clip,keepaspectratio]{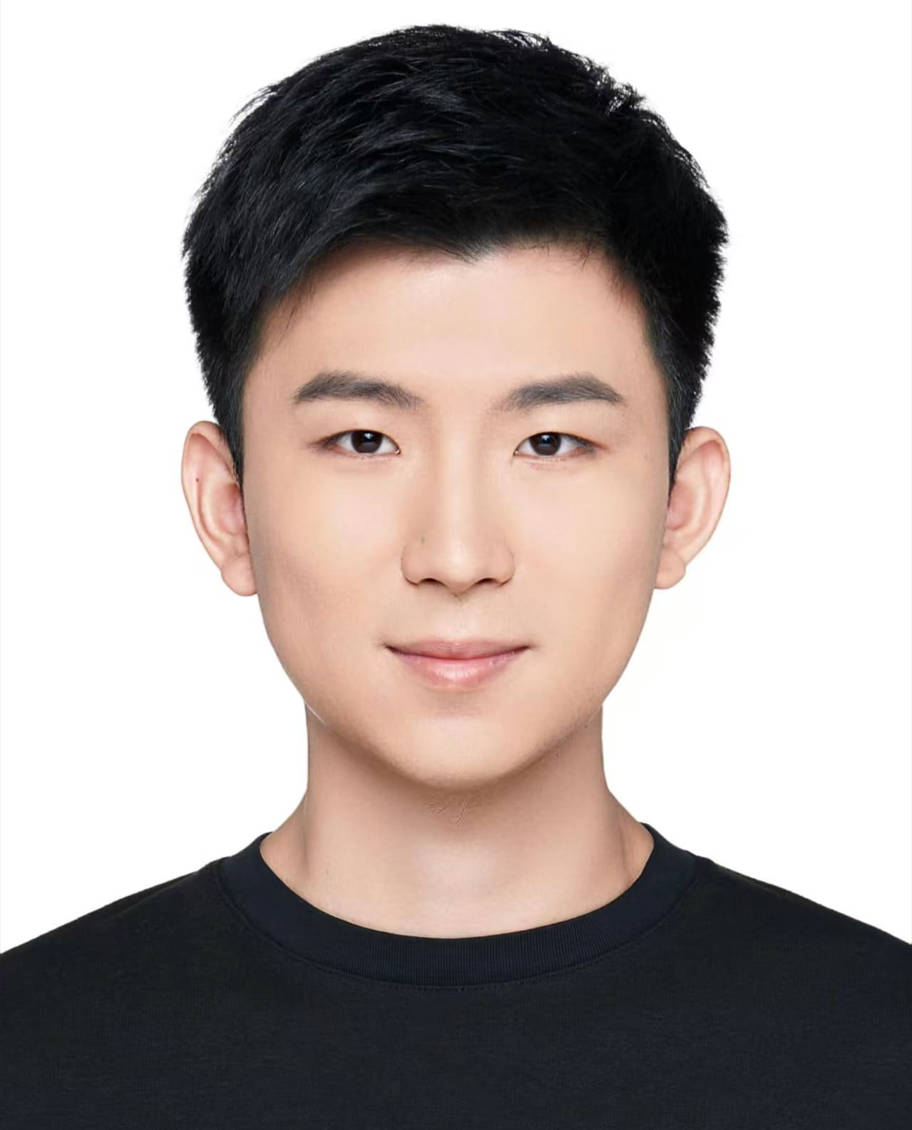}}]{Qijian Zhang} is a Senior AI Researcher in Tencent Games. He holds B.S. degree in Electronic Information Science and Technology from Beijing Normal University, Beijing, China (2019), and Ph.D. degree in Computer Science from City University of Hong Kong, Hong Kong SAR (2024). His current research interests include geometry processing, 3D generative models, and AI for games.
\end{IEEEbiography}

\vspace{-1.0cm}
\begin{IEEEbiography}[{\includegraphics[width=1in,height=1.25in,clip,keepaspectratio]{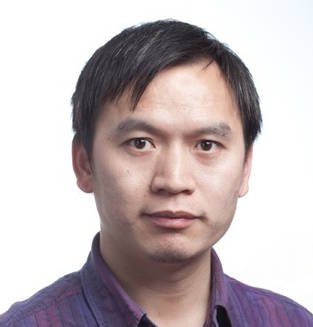}}]{Xiaozheng Jian} is an Expert Engineer and Principal Researcher in the TiMi L1 Studio of Tencent Games, Shenzhen, Guangdong, China. His expertise mainly lies in the fields of deep learning, generative models, and AI for games.
\end{IEEEbiography}

\vspace{-1.0cm}
\begin{IEEEbiography}[{\includegraphics[width=1in,height=1.25in,clip,keepaspectratio]{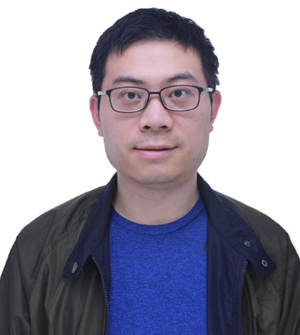}}]{Xuan Zhang} is a Principal Researcher in the TiMi L1 Studio of Tencent Games, Shenzhen, Guangdong, China. His research interests include deep learning, computer graphics, polygonal mesh processing, and AI-generated content.
\end{IEEEbiography}
\vspace{-0.8cm}
\begin{IEEEbiography}[{\includegraphics[width=1in,height=1.25in,clip,keepaspectratio]{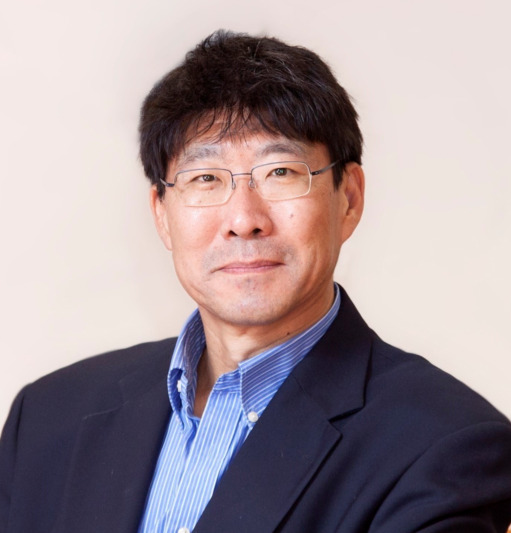}}]{Wenping Wang} (Fellow, IEEE) is a Professor of Computer Science and Engineering at Texas A\&M University. He conducts research in computer graphics, computer vision, scientific visualization, geometric computing, medical image processing, and robotics, and he has published over 400 papers in these fields. He received the John Gregory Memorial Award, Tosiyasu Kunii Award, and Bezier Award for contributions in geometric computing and shape modeling.
\end{IEEEbiography}

\vspace{-0.8cm}
\begin{IEEEbiography}[{\includegraphics[width=1in,height=1.25in,clip,keepaspectratio]{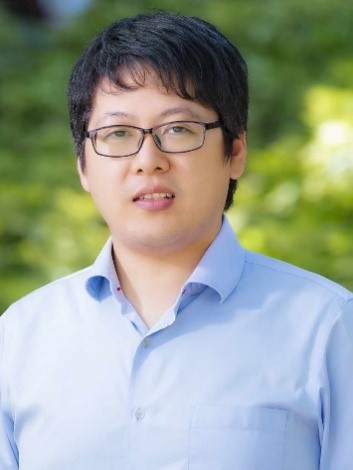}}]{Junhui Hou} (Senior Member, IEEE) is a Professor with the Department of Computer Science, City University of Hong Kong. His research interests are multi-dimensional visual computing.
	
Dr. Hou received the Early Career Award from the Hong Kong Research Grants Council, the Excellent Young Scientists Fund from the National Natural Science Foundation of China, and the IEEE SPS Best Paper Award. He is serving as a Senior Area Editor for \textit{IEEE Trans. on Image Processing} and an Associate Editor for \textit{IEEE Trans. on Visualization and Computer Graphics}, and \textit{IEEE Trans. on Multimedia}. He served as an Associate Editor for \textit{IEEE Trans. on Image Processing} and \textit{IEEE Trans. on Circuits and Systems for Video Technology}.
\end{IEEEbiography}

\vfill

\end{document}